\documentclass[10pt,conference]{IEEEtran}

\usepackage{cite}
\usepackage{microtype}
\usepackage{graphicx}
\usepackage{subcaption}
\usepackage{multirow}
\usepackage{tabu}
\usepackage{booktabs}
\usepackage[flushleft]{threeparttable} % for notes below table
\usepackage{multicol}
\usepackage{amsmath}
\usepackage{amsfonts}
\usepackage{xspace}
\usepackage{enumitem}
\usepackage{color}
\usepackage{xcolor}
\usepackage{soul}
\usepackage{extarrows}
\usepackage{stmaryrd}
\usepackage{gensymb}
\usepackage{lipsum}
\usepackage{tikz}
\usepackage{xcolor}
\usepackage{caption}
\usepackage{float}
\usepackage{todonotes}
\usepackage{stfloats} % position two-col figs
\usepackage[bottom]{footmisc}

\newcommand{\tablefontsize}[0]{
\fontsize{8pt}{10pt}\selectfont
}
\usepackage[ruled,linesnumbered]{algorithm2e}
\newlength\mylenin
\newcommand\myinput[1]{%
\settowidth\mylenin{\KwIn{}}%
\setlength\hangindent{\mylenin}%
\hspace*{\mylenin}#1\\}

\let\oldnl\nl% Store \nl in \oldnl
\newcommand{\nonl}{\renewcommand{\nl}{\let\nl\oldnl}}% Remove line number for one line

\newlength\mylenout

\makeatletter
\def\footnoterule{\relax%
  \kern-5pt
  \hbox to \columnwidth{\hfill\vrule width 1\columnwidth height 0.4pt\hfill}
  \kern4.6pt}
\makeatother

\newif\ifcomment

\commenttrue
\commentfalse 
            
\ifcomment
\definecolor{stelios_colour}{RGB}{191, 232, 255}
\newcommand{\stelios}[1]{\sethlcolor{stelios_colour}\hl{[Stelios: #1]}}

\definecolor{javier_colour}{RGB}{255, 204, 204}
\newcommand{\javier}[1]{\sethlcolor{javier_colour}\hl{[Javier: #1]}}

\newcommand{\blue}[1]{\textcolor{blue}{#1}}
\newcommand{\red}[1]{\textcolor{red}{#1}}
\else
\newcommand{\stelios}[1]{}
\newcommand{\javier}[1]{}
\newcommand{\blue}[1]{\textcolor{black}{#1}}
\newcommand{\red}[1]{\textcolor{black}{#1}}
\fi

\DeclareMathOperator*{\argmin}{argmin}

\def\BibTeX{{\rm B\kern-.05em{\sc i\kern-.025em b}\kern-.08em
    T\kern-.1667em\lower.7ex\hbox{E}\kern-.125emX}}
    
\newcommand{\tool}{unzipFPGA\xspace}

\newcommand\blfootnote[1]{%
  \begingroup
  \renewcommand\thefootnote{}\footnote{#1}%
  \addtocounter{footnote}{-1}%
  \endgroup
}

\usepackage{background}

\backgroundsetup{angle=0, 
scale=1,
color=black,
firstpage=true,
position=current page.north,
hshift=0pt,
vshift=-20pt,
contents={\ifnum\value{page}=1 PREPRINT: Accepted at the 29th IEEE International Symposium on Field-Programmable Custom Computing Machines (FCCM), 2021 \else \fi}
}

\begin{document}

\title{\vspace{-3mm}unzipFPGA: Enhancing FPGA-based CNN Engines with On-the-Fly Weights Generation \vspace{-0.3cm} %\vspace{-7mm}
}

\author{\IEEEauthorblockN{
Stylianos I. Venieris\IEEEauthorrefmark{2}\IEEEauthorrefmark{1},   
Javier Fernandez-Marques\IEEEauthorrefmark{3}\IEEEauthorrefmark{1},
Nicholas D. Lane\IEEEauthorrefmark{2}\IEEEauthorrefmark{4}
}
% \\
\IEEEauthorblockA{\IEEEauthorrefmark{2}Samsung AI Center, Cambridge, UK,
\IEEEauthorrefmark{3}University of Oxford,
\IEEEauthorrefmark{4}University of Cambridge}
\IEEEauthorblockA{Email: \{s.venieris, nic.lane\}@samsung.com, javier.fernandezmarques@cs.ox.ac.uk}
{\footnotesize \textit{\IEEEauthorrefmark{1}Indicates equal contribution.}}
\vspace{-0.6cm}
}

% ...
% \\
% \IEEEauthorblockA{\IEEEauthorrefmark{1}% 1st affiliations
% University of Oxford}
% \IEEEauthorblockA{\IEEEauthorrefmark{2}% 2nd affiliations
% Samsung AI Center, Cambridge, UK}
% \IEEEauthorblockA{\IEEEauthorrefmark{3} University of Cambridge}
% \IEEEauthorblockA{javier.fernandezmarques@cs.ox.ac.uk, \{s.venieris, nic.lane\}@samsung.com}
% }
	
\maketitle

\begin{abstract}

Single computation engines have become a popular design choice for FPGA-based convolutional neural networks (CNNs) enabling the deployment of diverse models without fabric reconfiguration. This flexibility, however, often comes with significantly reduced performance on memory-bound layers and resource underutilisation due to suboptimal mapping of certain layers on the engine's fixed configuration.
In this work, we investigate the implications in terms of CNN engine design for a class of models that introduce a pre-convolution stage to decompress the weights at run time. We refer to these approaches as \textit{on-the-fly}. 
To minimise the negative impact of limited bandwidth on memory-bound layers, we present a novel hardware component that enables the on-chip on-the-fly generation of weights. We further introduce an input selective processing element (PE) design that balances the load between PEs on suboptimally mapped layers. Finally, we present \textit{\tool}, a framework to train on-the-fly models and traverse the design space to select the highest performing CNN engine configuration. Quantitative evaluation shows that \tool yields an average speedup of 2.14$\times$ and 71\% over optimised status-quo and pruned CNN engines under constrained bandwidth \blue{and up to 3.69$\times$ higher performance density over the state-of-the-art FPGA-based CNN accelerators}.

% Single computation engines have become a popular design choice for FPGA-based convolutional neural networks (CNNs) enabling the deployment of models with diverse workloads while requiring no hardware-level reconfiguration. This flexibility, however, often comes with significantly reduced performance on memory-bound layers and resource underutilisation due to suboptimal mapping of certain layers on the engine's fixed configuration.
% In this work, we investigate the implications in terms of CNN engine design for a class of compressed models that introduce a pre-convolution stage to decompress the weights at run time. We refer to these approaches as \textit{on-the-fly}. 
% To minimise the negative impact of limited bandwidth on memory-bound layers, we present a novel hardware weights generator component as part of the CNN engine that enables the on-chip on-the-fly generation of weights. Furthermore, we introduce an input selective processing element (PE) design that balances the load between PEs on suboptimally mapped layers. Finally, we present \textit{\tool}, a framework to train on-the-fly models and traverse the %architectural 
% design space to select the highest performing CNN engine configuration. Quantitative evaluation shows that \tool yields an average speedup of 2.14$\times$ and 71\% over optimised status-quo and pruned CNN engines under constrained bandwidth, respectively, \blue{and up to 3.69$\times$ higher performance density over the state-of-the-art FPGA-based CNN accelerators}.

\end{abstract}

\vspace{-2mm}
\section{Introduction}
\label{sec:intro}
\vspace{-1mm}

% The emergence of convolutional neural networks (CNNs) as a core component in a variety of AI systems has led to the design of numerous FPGA-based accelerators.
% Currently, the landscape of CNN accelerators spans from CNN-specific processors~\cite{snowflake2017iscas,lightopu2020fpga,uniopu2020tvlsi} to highly tailored streaming architectures~\cite{streaming2016fpl,finn2017fpga,fpgaconvnet2019tnnls,lutnet2020toc}.
% One of the most widely adopted paradigms is the single computation engine~\cite{fpdnn2017fccm,dla2018fpl,cascadecnn2020date,caffeine2019tcad,dnnvm2019tcad,cascadecnn2018fpl,fft2020fpga,alamo2020tcad}. Under this approach, the accelerator comprises a powerful processing engine that is time-shared to execute the CNN layers sequentially. This approach enables the reuse of the accelerator resources across various CNNs and minimises the need for hardware-level reconfiguration upon deployment.

The emergence of convolutional neural networks (CNNs) as a core component in a variety of AI systems has led to the design of numerous FPGA-based accelerators.
Currently, the accelerator landscape spans from CNN-specific \mbox{processors~\cite{snowflake2017iscas,lightopu2020fpga,uniopu2020tvlsi}} to highly custom streaming architectures~\cite{streaming2016fpl,fpgaconvnet2019tnnls,finn2017fpga,lutnet2020toc}.
One of the most widely adopted paradigms is the single computation engine~\cite{fpdnn2017fccm,dla2018fpl,cascadecnn2020date,caffeine2019tcad,dnnvm2019tcad,cascadecnn2018fpl,fft2020fpga,alamo2020tcad,abdelfattah2020best,gamma2020iccad}, where a powerful processing engine is time-shared to execute the CNN layers sequentially. This approach enables the reuse of the accelerator resources across various CNNs and minimises the need for fabric reconfiguration upon deployment.

% Despite their flexibility, the performance of single computation engines is currently bounded by two main limitations: 
Despite its flexibility, performance is often bounded by:
1)~layers with low computation-to-communication ratio that become memory-bound~\cite{cascadecnn2018fpl,fft2020fpga,alamo2020tcad} and 2)~suboptimal mapping of diverse layers on the fixed engine configuration that leads to underutilised processing elements (PEs)~\cite{alamo2020tcad,latency2017fpl,cnnroofline2015fpga}. These two factors set a hard limit to the actual sustained performance that this family of accelerators can reach, indicating an emerging need for novel solutions to minimise their impact.

At the same time, a growing body of work, 
\red{orthogonal to other techniques such as pruning or quantisation,}
%from the ML community 
focuses on compressing CNNs through lossy non-structural methods~\mbox{\cite{ha2016hypernetworks, pmlr-v80-qiu18a, fernandez2018binarycmd, ijcai2018-380, ovsf2018emdl, Yang2020FSNet}}. % \red{aiming} to reduce their high memory burden. 
Under this new paradigm, the weights of a model are deployed in a compact form and are ``inflated" at run time.
Given that several CNN layers are constrained by the limited off-chip memory
bandwidth~\cite{latency2017fpl,cascadecnn2018fpl,fft2020fpga,alamo2020tcad}, storing the compressed weights on-chip and reconstructing them on-the-fly can play a key role in alleviating the memory-boundedness and enabling better utilisation of the computational resources.
Nevertheless, the novel dataflow and execution scheme 
of such models brings up a new challenge regarding their optimised mapping.
Existing accelerators have been designed for conventional deep models, adopting either a streaming or layer-by-layer execution~\cite{cnnfpgatoolflows2018csur}.
Hence, despite the significant potential of on-the-fly models, their different execution paradigm renders conventional architectures futile in serving them.

In this work, we present \tool, a novel CNN inference system that alleviates the withstanding limitations of single computation engines. % for CNN inference. 
To minimise the impact of memory-bound layers, \tool first introduces a mechanism for the on-the-fly generation of weights, reducing transfers from the off-chip memory. Moreover, we propose an input selective PE design that counteracts the underutilisation of the computational resources due to the suboptimal mapping of diverse layers.
Overall, this work makes the following key contributions:

\begin{itemize}
    
    \item A novel CNN hardware architecture for on-the-fly generation of CNN weights.
    \red{We introduce a weights generator with custom memory organisation and a highly optimised datapath that enables sustaining high processing rates.}
    The configurable parameters are exposed to the system-level design space exploration to yield the highest performing allocation of resources between the weights generator and the core CNN engine \blue{for the given CNN-FPGA pair}.

    \item A new CNN engine design comprising an array of input selective PEs. A subset of PEs is enhanced with efficient switches that enable other PEs to perform work stealing. 
    \red{In this manner, the proposed design applies workload balancing, improving up to 20\% the performance on layers that previously severely underutilised the instantiated PEs.}

    \item
    \red{A framework for deriving and mapping on-the-fly models from pretrained CNNs on a given FPGA,}
    {yielding 2.48$\times$ average %(2.14$\times$ geo. mean)
    performance density gain over various state-of-the-art FPGA designs. Compared to a state-of-the-art pruning framework~\cite{Molchanov_2019}, \tool delivers up to 90\% and 69\% compression and throughput gain under the same accuracy.
    % while delivering the same accuracy
    }
\end{itemize}

% \vspace{-0.6em}
\vspace{-3mm}
\section{Background \& Related Work}
\label{sec:background}
% \vspace{-0.2em}
\vspace{-1mm}

\subsection{Designing Lightweight Convolutional Neural Networks}
\vspace{-0.2em}
\red{
The plethora of existing techniques to modify CNNs for faster inference can be categorised into: pruning~\cite{luo2017thinet, blalock2020state, he2018soft}, which removes redundant parameters; % and often leads to fewer operations; 
quantisation~\cite{krishnamoorthi2018quantizing,jacob2018quantization,dong2019hawq,fernandezmarques2020searching,alizadeh2018a}, which results in low-precision compact models; or, sparsification~\cite{wen2016learning, sparsify2016sensys, gale2020sparse}, which leverages compressed data formats. In addition, a number of frameworks combine several of these techniques. Most notably: Deep Compression~\cite{deepcompression2015iclr} which, given an over-parametrised model, 
applies pruning, quantisation and Huffman encoding;  RedCNN~\cite{pmlr-v70-wang17m} which prunes channels based on an activation overlap metric; %, reducing latency and memory usage; 
and, more recently, APQ~\cite{wang2020apq}, which designs a CNN that meets given compute, memory and latency constraints through a joint optimisation formulation.
}

\vspace{-0.3em}
\subsection{On-the-Fly Convolutional Neural Networks}
\vspace{-0.2em}

Orthogonal to \red{these methods}, various works have explored ways of \red{factorising} the filters in CNNs in order to produce compressed model representations. Common to these techniques is the need for a \textit{decompressing} stage that generates the filters \textit{on-the-fly} during inference. 
A selection of such %these on-the-fly 
techniques include: \cite{ha2016hypernetworks} that uses an auxiliary NN to generate each layer's weights in the main network given an embedding of the weights.  
In~\cite{pmlr-v80-qiu18a}, weight filters are constructed as a dense combination of a set of Fourier Bessels bases that are generated deterministically at run time.
Another technique exploiting deterministic bases was presented in~\cite{fernandez2018binarycmd,ijcai2018-380, ovsf2018emdl}, where bases are formed from orthogonal variable spreading factor (OVSF) binary codes. 
\red{It enables the construction of model weights} by learning a linear combination of OVSF bases during training. Finally, in~\cite{Yang2020FSNet} 
% FSNet~\cite{Yang2020FSNet} 
the filters of each layer are sampled from a single learnable filter, inducing weight sharing across filters. This sampling process is deterministic and enables faster processing via an associated integral image-based implementation.

In this work, we make use of OVSF codes to compress filters in a CNN, directly reducing the overheads due to off-chip memory accesses to retrieve weights, based on three aspects: 
1)~these codes are \textit{binary} and thus can be efficiently stored on-chip~\cite{5291322}; 2)~their theoretical properties are well studied by the wireless community~\cite{Andreev:2003:OCG:764808.764868,1411169}; 3)~they offer good compression-accuracy trade-off in various \blue{AI} tasks~\cite{ovsf2018emdl,ijcai2018-380}.
\blue{
Nonetheless, \cite{ovcf-fpga}~is the sole existing FPGA-based OVSF design, presenting a direct implementation specific for communication systems. To effectively use OVSF with CNNs, the underlying design needs to be tailored and optimised for the CNN dataflow.
}

\vspace{-0.3em}
\subsection{On-the-Fly OVSF Models}
\label{sec:ontheflyOVSFmodels}
\vspace{-0.2em}

In this section, we give an overview on how OVSF codes are generated and their use for constructing CNN weights.

\subsubsection{Constructing OVSF bases}

OVSF codes can be recursively constructed as an $n$-step expansion of an $L\times L$ Hadamard matrix, $H_{n}$, with $L = 2^{n}$, $n\in \mathbb{N}$. \red{The resulting $H_{n}$ is comprised of $L$ mutually orthogonal binary codes.}

The length, $L$, of the OVSF codes depends on the shape of the tensor that is generated, but also on the granularity at which this process happens.
\red{In this work, we construct $N_{\text{in}}\times$$K\times$$K$ filters by concatenating $N_{\text{in}}$ $K$$\times$$K$ matrices, requiring the generation of $L$$=$$K^{2}$ OVSF codes. Alternative approaches exist~\cite{ovsf2018emdl}, each with different compute and memory overheads.}
% For instance, a brute-force method of generating a $N_{\text{in}}\times$$K\times$$K$ filter requires codes with $L$$=$$N_{\text{in}}K^{2}$. However, that same filter can be instead constructed by concatenating $N_{\text{in}}$ $K$$\times$$K$ matrices, requiring the generation of $L$$=$$K^{2}$ codes. Each approach has different implications in terms of memory and compute overheads~\cite{ovsf2018emdl}. In this work, we use the latter since, by excluding $N_{\text{in}}$ from the bases generation, it allows for this parameter to be any natural number. 

% \vspace{-0.3em}
\subsubsection{Generating CNN filters with OVSF bases}
\label{sec:ovsf_lin_comb}
% \vspace{-0.2em}

A $N_{\text{in}}$$\times$$K$$\times$$K$ weight filter, $f_{i}$, in a convolutional layer can be represented as a truncated linear combination of OVSF codes. The codes are first reshaped to match the shape of $f_{i}$. Mathematically, this linear combination can be represented as $f_{i} = \sum_{j = 0}^{\lfloor \rho \cdot l \rfloor}\alpha_{i}^{j}B_{i}^{j}$, where $\rho \in [0,1]$ is the ratio of codes to use in order to generate filter $f_{i}$, $l$ the total number of OVSF codes of length \mbox{$L=$$K^{2}N_{\text{in}}$}, $B_{i}^{j}$ the j-th OVSF code and $\alpha_{i}^{j} \in \mathbb{R}$ its associated scalar. %Throughout this work we construct $f_{i}$ as a concatenation of $N_{\text{in}}$ $K$$\times$$K$ bases, resulting in $L$$=$$K^{2}$ and allowing for filters with any number of channels to be constructed with OVSF.
%Recall that because $l$ codes of length $L$ form a basis, then $l=L$. This also implies that 
When $\rho$$=$$1$, the number of scalars, $|\alpha_{i}|$, is equal to the number of weights in $f_{i}$. These scalars are learnt via standard backpropagation. A compressed representation of $f_{i}$ is obtained when $\rho$$<$$1$.
%The accuracy degradation due to compression can be controlled with careful fine-tuning and hyperparameter selection.
Upon deployment, the filters are first generated and then the main inference computation proceeds as normal.

% One limitation of OVSF codes comes in the form of only being able to form a basis of $\mathbb{R}^{L}$ with $L$ being a power of two (\textit{e.g.} $N_{\text{in}}$$\times$$4$$\times$$4$). This complicates the usage of OVSF to generate $3$$\times$$3$ filters, which are popular in CNN models. %This limits OVSF-based filters to be of shape $C$$\times$$2$$\times$$2$, $C$$\times$$4$$\times$$4$, etc.
%In Section~\ref{sec:OVSFmodels}, we present \tool's technique for overcoming this limitation by extracting $3$$\times$$3$ from  $4$$\times$$4$ filters.
\red{By design, OVSF codes cannot directly be used to generate $3$$\times$$3$ filters, which are popular in CNN models.}
In Section~\ref{sec:OVSFmodels}, we present a technique to overcome this limitation.
% filters from the $4$$\times$$4$ filters generated with OVSF codes. 

% \newpage
\vspace{-0.3em}
\subsection{Challenges of FPGA-based CNN Inference Engines}
\label{sec:cnn_accelerators}
\vspace{-0.2em}

Until now, a wide array of FPGA-based CNN accelerators have been proposed. 
% In the customisation-programmability spectrum, 
Existing designs span from custom streaming architectures~\cite{streaming2016fpl,fpgaconvnet2019tnnls} and accelerators for quantised~\cite{finn2017fpga,ternaryfpga2018trets,finnr2018trets,multiprec2019fpt,lutnet2020toc,logicnets2020fpl} and sparse CNNs~\cite{cambriconx2016micro,cambircons2018micro,scnn2017isca,sparten2019micro,sparsecnnaccel2019fccm,extensor2019micro,tensaurus2020hpca}, up to instruction-based processors~\cite{snowflake2017iscas,lightopu2020fpga,uniopu2020tvlsi}.
One of the most well adopted paradigms are the single computation engines~\cite{fpdnn2017fccm,dla2018fpl,cascadecnn2020date,caffeine2019tcad,dnnvm2019tcad,cascadecnn2018fpl,fft2020fpga,alamo2020tcad,abdelfattah2020best}, due to their balanced trade-off of programmability and performance. 
Currently, \blue{despite the progress in processing unit design, further gain in the attainable performance of such engines is hindered by two main factors:}
i)~memory-bound layers that are dominated by the communication with the external memory~\cite{mem_req2918iiswc,cascadecnn2018fpl,fft2020fpga,alamo2020tcad}. While embedded platforms provide limited bandwidth~\cite{lostbw2019fpt,divrsemem2019fpt,venieris2018fpl}, \textit{e.g.} less than 4.5~GB/s for Ultra96 and ZC706, sustaining peak bandwidth even on larger devices, such as ZCU104, is nontrivial~\cite{divrsemem2019fpt}. This is aggravated as multiple applications are collocated on a single device~\cite{venieris2018fpl,codesignmultiple2020dac,multinn2020isca}; and ii)~underutilised PEs due to the mismatch of diverse layer shapes~\cite{alamo2020tcad,latency2017fpl,cnnroofline2015fpga,gamma2020iccad,abdelfattah2020best}.

\blue{
\textbf{Memory-centric Designs.}
The memory bandwidth problem faced by CNN engines has been studied in previous work. 
EIE~\cite{eie2016isca} uses the Deep Compression method~\cite{deepcompression2015iclr} to compress FC layers' weights. However, as FC layers have been mostly abandoned in modern CNNs, its applicability is limited.
Angel-Eye~\cite{angeleye2018tcad} compresses all layers through precision quantisation. 
Cambricon-X~\cite{cambriconx2016micro} transfers only the non-zero weights, while Cambricon-S~\cite{cambircons2018micro} and Scalpel~\cite{scalpel2017isca} apply coarse weight pruning, but with significant accuracy drop.
CircCNN~\cite{circcnn2017micro} uses block-circulant matrices for weights, but requires complex FFT hardware. % for efficient execution.
\cite{permdnn2018micro} uses permuted diagonal matrices for sparse weights, but only targets FC layers. 
% converts sparse weights to permuted diagonal matrices, but only focuses on FC layers.
\cite{escher2017fccm}~exploits large batch sizes to increase weights reuse and, thus, is not suitable for latency-critical applications that cannot tolerate batching~\cite{latency2017fpl}.
}

\blue{
Focusing on activations, \cite{fusedlayer2016micro} fuses adjacent layers to cache intermediate activations, while Eyeriss~\cite{eyeriss2017jssc}, \cite{def2021aspdac} and~\cite{scnn2017isca} employ encoding schemes to minimise their bandwidth footprint.
Other solutions have either relied on large devices~\cite{opencldla2017fpga} and %multi-FPGA setups
multiple FPGAs~\cite{brainwave2018isca} to fit all weights on-chip, or utilised highly customised designs to exploit multi-precision cascades~\cite{cascadecnn2020date} or fine-grained pruning~\cite{sparsecnnaccel2019fccm} at the cost of notable accuracy drop.
}

\blue{
\textbf{Tackling PE Underutilisation.}
So far, a limited number of designs have focused on ii). \cite{maximising2017isca} 
% addresses underutilisation by grouping 
groups CONV layers based on the compatibility of their shapes. % dimensions. 
\cite{streaming2016fpl} maps each layer to a dedicated compute stage, which can be used only for shallower networks, % (\textit{e.g.} AlexNet, VGG16), 
but does not scale to the deeper models of today. Furthermore, a limited number of works rely on FFT-based designs with flexibile dataflow~\cite{fft2020fpga} and costly ASIC solutions~\cite{flexflow2017hpca,recpatterns2017tvlsi,maeri2018asplos} with highly flexible PE interconnect.
}

In contrast to these works, we propose an approach that is independent of the CNN engine by not requiring any modification to the engine architecture itself. \tool can benefit any existing single computation engine by augmenting it with its hardware weights generator and enhancing its PE array with lightweight switches, without affecting the PE's internal processing units. 
% (\textit{e.g.}~dot-product circuit or MAC operators). 
\blue{
As such, \tool is orthogonal and complementary to quantisation~\cite{angeleye2018tcad}, activations' encoding~\cite{eyeriss2017jssc,scnn2017isca}, fusion~\cite{fusedlayer2016micro} and zero-skipping PEs~\cite{cnvlutin2016isca,pragmatic2017micro,cambircons2018micro,shapeshifter2019micro,sparten2019micro}.
}

\vspace{-0.2em}
% \clearpage
\section{\tool}
\vspace{-0.35em}

\vspace{-0.3em}
\subsection{Overview}
\label{sec:overview}
\vspace{-0.2em}

To remedy the limitations of existing FPGA-based CNN engines, \tool employs a weights generation scheme that minimises the external memory bandwidth requirements. The proposed system introduces a hardware weights generator based on OVSF bases 
that enables the on-the-fly, on-chip generation of weights, significantly reducing the off-chip memory accesses. To alleviate the underutilisation of PEs due to layer shape mismatch, \tool introduces input selective PEs, enabling seamless load balancing with minimal hardware modifications.

\begin{figure}[t]
    \centering
    \vspace{-0.3cm}
    \includegraphics[width=0.8\columnwidth,trim={3cm 7.5cm 12.25cm 0cm},clip]{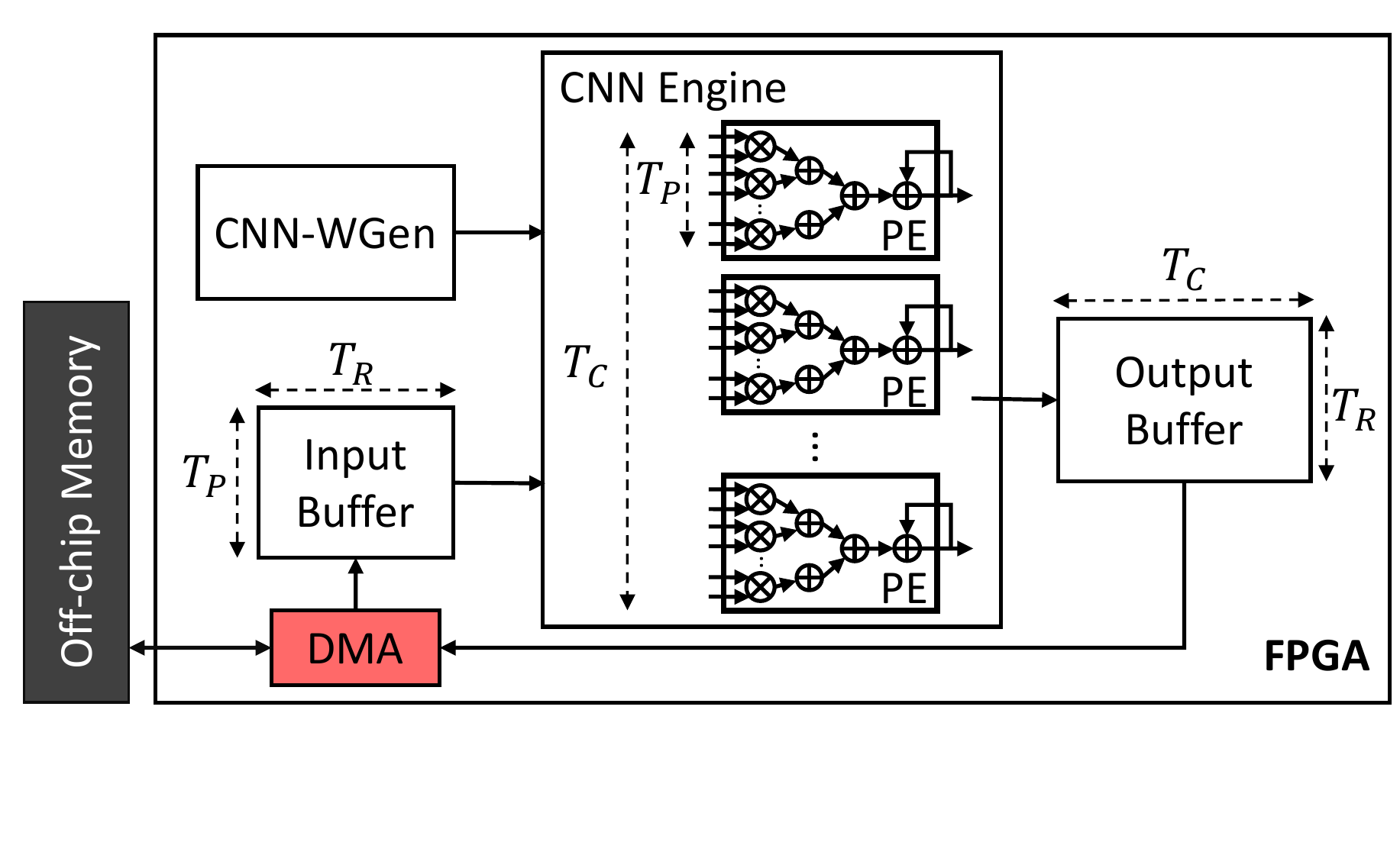}
    \vspace{-0.95cm}
    \caption{\footnotesize Overview of \tool{}'s architecture. 
    }
    % \vspace{-0.4cm}
    \vspace{-2em}
    \label{fig:high_level_arch}
\end{figure}

The overall flow of the proposed framework is as follows.
First, given a CNN, \tool derives an OVSF variant, by replacing the CONV weights with a trainable linear combination of OVSF bases and selecting each layer's compression ratio $\rho$ (Sec.~\ref{sec:ontheflyOVSFmodels}). Next, the OVSF model is trained using the supplied training set (Sec.~\ref{sec:OVSFmodels}). The OVSF model is passed to the design space exploration (DSE) phase to explore different resource allocations between the weights generator (\texttt{CNN-WGen}) and the CNN engine (Fig.~\ref{fig:high_level_arch}) (Sec.~\ref{sec:dse}). Upon completion, the DSE yields the highest performing configuration of \tool's architecture (Sec.~\ref{sec:arch}) tailored to the given CNN-device pair and the system is deployed on the target FPGA platform.

% \vspace{-0.15cm}
\vspace{-0.3em}
\subsection{Architecture}
\label{sec:arch}
% \vspace{-0.05cm}
\vspace{-0.2em}

Fig.~\ref{fig:high_level_arch} shows the proposed architecture consisting of three components: the CNN hardware weights generator (\texttt{CNN-WGen}), the core CNN engine and the I/O subsystem. From an operational perspective, the layers of the target CNN are scheduled sequentially, 
% over the accelerator, 
with the computation of the current layer 
% by the CNN engine 
and the I/O communication of adjacent layers overlapped in a pipelined manner. 
A key component of the proposed architecture is the \texttt{CNN-WGen} module, which relaxes the off-chip memory bandwidth requirements by generating the weights % of each layer 
at run time with minimal access to the external memory.

\vspace{-0.3em}
\subsection{CNN Engine}
\label{sec:cnn_engine}
\vspace{-0.2em}

To execute various layer shapes and types, the core CNN engine comprises a parametrised array of PEs for the execution of block general matrix multiply (GEMM). %(Fig.~\ref{fig:high_level_arch}). 
By interpreting convolutions as matrix multiplication, both CONV and FC layers are supported. % by the CNN engine. 
In this manner, a CONV layer with $N_\text{in}$ \blue{$H$$\times$$W$} input activations, $N_\text{out}$ output channels, $K$$\times$$K$ filters, $p$ padding and $S$ stride involves the multiplication between an $R$$\times$$P$ activations matrix and a $P$$\times$$C$ weights matrix in order to produce a $R$$\times$$C$ output matrix, with $R$$=$$\left\lceil \frac{H + 2p - K}{S}+1 \right\rceil \left\lceil \frac{W + 2p - K}{S} + 1  \right\rceil$, $P$$=$$N_{\text{in}} K^2 $ and $C$$=$$N_{\text{out}}$.
The engine is parametrised with respect to the tile size $\left<T_R,T_P,T_C\right>$ of each matrix dimension $\left<R,P,C\right>$, the number of PEs and the MAC operators within each PE.

\begin{figure}[t]
    \centering
    \vspace{-0.36cm}
    {
    \includegraphics[width=0.85\columnwidth,trim={4cm 12.5cm 19cm 2.7cm},clip]{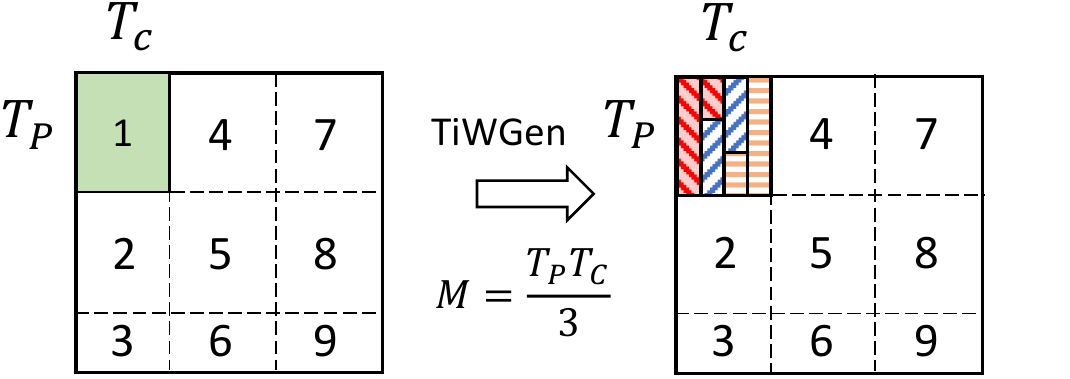}
    }
    \vspace{-0.1cm}
    \caption{\footnotesize An example of the TiWGen technique. With a tile size of $M$, each tile is generated in $\lceil T_P T_C/M \rceil \cdot \rho K^2$ cycles.}
    % \vspace{-0.005cm}
    \label{fig:weights_gen_tiling}
\end{figure}

To produce a {\small $T_R$$\times$$T_C$} output tile, {\small $\left\lceil \frac{P}{T_P}\right\rceil$} tiles from the activations and weights matrices are processed and accumulated sequentially. 
\tool's mapping strategy (Fig.~\ref{fig:high_level_arch}) ties $T_P$ to the MAC units per PE to exploit the parallelism within each $T_P$-wide dot product, and $T_C$ to PEs to parallelise the dot products at each output column. Overall, the rows of the {\small $T_R$$\times$$T_P$} activations tile are processed in a pipelined manner to maximise throughput.
\blue{This is equivalent to an output stationary dataflow~\cite{shidiannao2915isca,eyeriss2017jssc}, which minimises the memory accesses for the output activations by caching partial sums on-chip. 
Nonetheless, \tool is adaptable to other dataflows.
}

\vspace{-1em}
\subsection{CNN Hardware Weights Generator}
\label{sec:hw_weights_gen}
\vspace{-0.3em}

To attain high performance, the adopted weights generation algorithm is mapped to hardware via two key techniques: a \textit{tiled weights generation} (TiWGen) method and a hardware weights generator (\texttt{CNN-WGen}).

\subsubsection{Tiled Weights Generation}
\label{sec:tiled_weights_gen}

To be able to target layers of various dimensions, we introduce a tiling method for the weights generation process, denoted by TiWGen. As shown in Fig.~\ref{fig:weights_gen_tiling}, TiWGen divides each $T_P$$\times$$T_C$ weights tile into subtiles of size $M$, with $M$ being uniform across the CNN's layers.
Tiling on top of the weights generation method makes the data flow of diverse layers identical to each other.
% At run time, the OVSF basis vectors are blocked into fixed shapes with a uniform tile size $M$ across the CNN's layers.
With this approach, the value of $M$ becomes independent of the CNN model and is solely bound by the on-chip computational resources allocated to the weights generator. Thus, $M$ provides a tunable trade-off between weights generation speed and resource consumption.

Alg.~\ref{alg:tiled_weights_gen} describes the overall TiWGen technique. Each $T_P$$\times$$T_C$ tile of the weights matrix is processed sequentially~(line 1). This is done by partitioning each tile into $\left\lceil \frac{T_P T_C}{M} \right\rceil$ subtiles~(line 2). After all basis vectors of the current subtile have been processed~(lines 4-9), the associated part of the output tile is updated and the algorithm proceeds to the next subtile. When all subtiles of a tile have been generated, the weights matrix is updated (line 12) and the algorithm continues to the next iteration until all weights tiles have been formed.

\SetArgSty{textnormal} % normal text in if-conditions
\setlength{\textfloatsep}{0pt}% Remove \textfloatsep
\begin{algorithm}[!t]	
%	\algsetup{linenosize=\tiny}
	\footnotesize
%	\scriptsize
	\SetAlgoLined
	\LinesNumbered
	\DontPrintSemicolon
	
	% inputs
	\KwIn{Layer's weights matrix shape:  $P \times C = N_{\text{in}}K^2 \times N_{\text{out}}$}
	\nonl
	\myinput{Row and column tile sizes $T_P$ and $T_C$}
	\nonl
	\myinput{$\alpha$ values with $\alpha \in \mathbb{R}^{N_\text{in}N_\text{out}\left\lceil \rho K^2 \right\rceil}$}
	
	% outputs
	\KwOut{Weights matrix $\boldsymbol{W}$}
	
	\For(// \textit{tiles loop} - \textbf{\#} \textbf{\texttt{PIPELINE}}){$t\gets1$ \KwTo $\left\lceil \frac{P}{T_P} \right\rceil$$\cdot$$\left\lceil \frac{C}{T_C} \right\rceil$}{
	   % \textbf{\#} \textbf{\texttt{CNN-WGen}}: \textit{pipeline}\;
    	\For(// \textit{subtiles loop} - \textbf{\#} \textbf{\texttt{PIPELINE}}){$i\gets1$ \KwTo $\left\lceil \frac{T_P T_C}{M} \right\rceil$}{
    	   % \textbf{\#} \textbf{\texttt{CNN-WGen}}: \textit{pipeline}\;
    	    $\text{subtile}^t_i \leftarrow \mathbf{0}$ \;
    	    \For(// \textit{basis vectors loop} - \textbf{\#} \textbf{\texttt{PIPELINE}}){$j\gets1$ \KwTo $\rho K^2$}{
    	       % \textbf{\#} \textbf{\texttt{CNN-WGen}}: \textit{pipeline}\;
    	        \For(// \textbf{\#} \textbf{\texttt{UNROLL}}){$k\gets1$ \KwTo $M$}{
    	           % \textbf{\#} \textbf{\texttt{CNN-WGen}}: \textit{unroll}\;
    	            $incr_k \leftarrow \text{vec}_j(k) \cdot \alpha_k$ // \textit{Multiplier array} \;
    	            $\text{subtile}^t_i(k) \leftarrow \text{subtile}^t_i(k) + incr_k$ // \textit{Adder array}\;
    	        }
    	    }
    	    $\text{tile}^t \leftarrow \text{UpdateTile}(\text{tile}^t,\text{subtile}^t_i)$\;
    	}
    	$\boldsymbol{W} \leftarrow \text{UpdateMatrix}(\boldsymbol{W}, \text{tile}^t)$\;
	}

	\caption{\small 
	Generation of a layer's weights using TiWGen
	}
	\label{alg:tiled_weights_gen}	
\end{algorithm}

\subsubsection{Microarchitecture}
\begin{figure}[t]
    \centering
    \vspace{-0.1cm}
    % \fbox
    {
    \includegraphics[width=1\columnwidth,trim={0.45cm 0cm 2.8cm 0cm},clip]{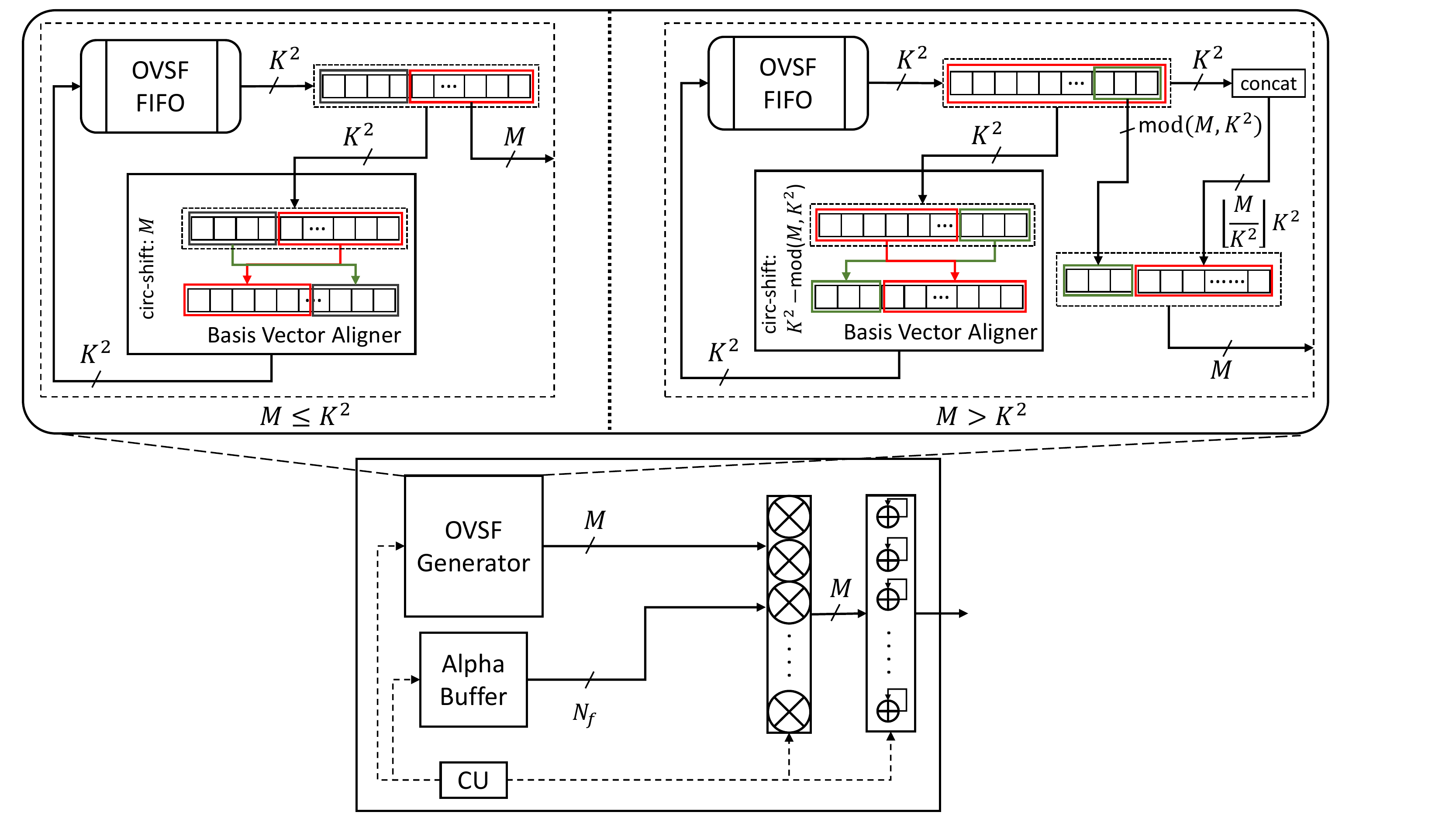}
    }
    \vspace{-0.4cm}
    \caption{\footnotesize Microarchitecture of the \texttt{CNN-WGen} module.}
    % \vspace{-0.2cm}
    \label{fig:hw_weights_gen}
\end{figure}

Key enabler to the proposed design is the \texttt{CNN-WGen} module, which is responsible for generating the weights in an orderly manner and feeding them to the CNN engine. 
Its main components comprise (Fig.~\ref{fig:hw_weights_gen}):
i)~a compute datapath formed by two vector %arithmetic 
units (multiplier and adder arrays), ii)~the Alpha buffer storing the $\alpha$ values, and iii)~the OVSF generator that is responsible for outputting the $M$-sized basis vector subtiles as dictated by the TiWGen scheme. 

\texttt{CNN-WGen}'s strategy for mapping the TiWGen method to hardware is annotated in Alg.~\ref{alg:tiled_weights_gen}.
\texttt{CNN-WGen} implements TiWGen by pipelining the three outer loops over tiles, subtiles and basis vectors, and unrolling the inner loop that processes the $M$-sized subtile.
To unroll the inner loop, \texttt{CNN-WGen} employs two $M$-wide vector units that perform $M$-parallel multiplications and additions, respectively. In this manner, tuning $M$ can balance the parallelism-resource usage trade-off.

\textbf{Compute Datapath.}
With reference to Fig.~\ref{fig:hw_weights_gen}, the vector arithmetic units %(\textit{i.e.}~the multiplier and adder arrays) 
must have a fixed number of inputs that meets the resource budget of the target FPGA and namely the available DSPs. % blocks. 
For the i-th subtile (line 3 in Alg.~\ref{alg:tiled_weights_gen}), $\rho K^2$ vectors of size $M$ are produced by the OVSF generator and the associated $\alpha$ values fetched from the Alpha buffer, and are fed to the multiplier array in a pipelined manner. All $M$ elements are processed in parallel by the $M$-wide vector units, leading to the unrolling of the inner loop on line 5 of Alg.~\ref{alg:tiled_weights_gen}. The adder array processes the outputs of the multiplier array by accumulating the $\rho K^2$ intermediate results. Finally, when the processing proceeds to the next subtile (\textit{i.e.}~next iteration of the loop on line 2), the control unit (CU in Fig.~\ref{fig:hw_weights_gen}) resets the accumulators' state. Overall, the design-time configurable parameter $M$ determines the size of the vector units and controls in this way the performance-resource trade-off of \texttt{CNN-WGen}. The tuning of $M$ is exposed to the DSE (Sec.~\ref{sec:dse}) %in order 
to determine how many resources are allocated to \texttt{CNN-WGen}.

\textbf{Alpha Buffer.}
Following TiWGen, each %\mbox{$M$-sized} 
subtile contains weights from $N_f$ distinct $K$$\times$$K$ filters (Eq.~(\ref{eq:filters_per_subtile})). To sustain the throughput of \texttt{CNN-WGen}, an equal number of $\alpha$s has to be fetched in parallel from the Alpha buffer.
This is accomplished by designing the buffer with appropriate memory organisation and addressing. Each layer contains $N_{\text{in}} N_{\text{out}} \left\lceil \rho_l K_l^2 \right\rceil$ distinct $\alpha$ values. The RAM blocks are organised with $N_P^{\text{Alpha}}$$=$$N_f$ ports %(Eq.~(\ref{eq:alpha_buff_ports})) 
%that provide a bandwidth of $N_f$ $\alpha$s per cycle, 
and a depth of $D^{\text{Alpha}}$ (Eq.~(\ref{eq:alpha_buff_depth})) to accommodate $N_L$ layers.
\vspace{-1mm}
\begin{equation}
    \footnotesize
% \begin{align}
    \resizebox{0.725\linewidth}{!}{
    $N_{f} = \left\lceil \frac{\min (T_P,M)}{K_\text{max}^2} \right\rceil \left\lfloor \frac{M}{T_P} \right\rfloor + \text{mod}(M,T_P) \left\lceil \frac{M}{K_\text{max}^2} \right\rceil$
    }
    \vspace{-1mm}
    \label{eq:filters_per_subtile}
\end{equation}
\begin{equation}
    \footnotesize
    \resizebox{0.725\linewidth}{!}{
    $D^{\text{Alpha}} = \overbrace{\sum_{l=1}^{N_L}}^{\text{for each layer}} \frac{\overbrace{N_{\text{in}}^l N_{\text{out}}^l \left\lceil \rho_l K_l^2 \right\rceil}^{\text{no. of $\alpha$ values}}}{N_P^{\text{Alpha}}} \quad \text{ ( \textit{Buffer depth} )}$
    }
    \label{eq:alpha_buff_depth}
    \vspace{-0.1cm}
    \vspace{-0.2cm}
\end{equation}
where $N_L$ is the number of layers, $N_{\{\text{in},\text{out}\}}^l$ the l-th layer's number of input/output channels and $\rho_l$ the compression ratio.

\textbf{OVSF Generator.}
Based on TiWGen, the basis vectors are processed in a blocked manner with a tile size of $M$. % as specified by Algorithm~\ref{alg:tiled_weights_gen}. 
This approach leads to two pipelined loops over the $\left\lceil \frac{T_P T_C}{M} \right\rceil$ subtiles (line 2) and the $\rho K^2$ basis vectors (line 4) and the unrolled loop of processing the $M$-element subtile with the vector units (line 5). To produce the i-th subtile, the OVSF generator has to feed the compute datapath with $\rho K^2$ basis vectors that are tiled as dictated by TiWGen's parameter $M$.

In order not to slow down the operation of \texttt{CNN-WGen}, the OVSF generator has to match the %processing 
rate of the vector units by providing %a bandwidth of 
$M$~bits/cycle.
A conventional approach % of achieving this 
involves laying out the tiled vectors in an on-chip buffer through static data marshalling so that the %number of 
ports match the required rate (\textit{i.e.}~%\mbox{$M$ bits/cycle $\mapsto$ 
$M$~ports) with a depth equal to the number of reads/tile (\textit{i.e.}~\mbox{\#basis vectors$\times$\#subtiles}).
However, such a design would replicate %lead to the replication of 
the basis vectors either in the same address %through concatenation 
(\textit{e.g.}~when $M$$>$$K^2$) or in multiple addresses (\textit{e.g.} storing rotated versions %of the vectors 
as required by different subtiles). This %design 
leads to inefficient utilisation of on-chip memory due to significant replication. 

Another approach that would avoid the basis vector replication involves the instantiation of a $K^2$-deep OVSF memory with each location storing one $K^2$-bit vector. Such a design requires significantly lower amount of storage and provides an access rate of 1 vector/cycle by reading from the appropriate address. However, to obtain the $M$-bit subtile from the $K^2$-bit vector, complex multiplexer selection circuitry has to be instantiated. % that would dynamically construct the subtile. 
This approach can affect the attainable clock frequency or add latency cycles and, thus, degrade \texttt{CNN-WGen}'s performance.

To alleviate these limitations when mapping TiWGen's tiling scheme, 
a custom OVSF generator was developed \mbox{(Fig.~\ref{fig:hw_weights_gen} - top)}. By utilising a FIFO for the OVSF vectors together with a basis vector aligner, the OVSF generator introduces a rate-matching mechanism that sustains the processing rate of the compute datapath while efficiently utilising the on-chip memory. 
\blue{The generator performs a different operation depending on the values of $M$ \mbox{and $K_i^2$} of layer $i$ (Fig.~\ref{fig:hw_weights_gen}).
}

Initially, %in both \blue{flows}, 
the OVSF FIFO stores the \mbox{{\small ($K_i^2K_i^2$)}-bit} basis vectors. %\footnote{Based on OVSF theory, an $\mathbb{R}^{K^2}$ base requires $K^2$ components.} 
The current vector is read from the FIFO into the top register. 
If {\small $M$$\le$$K_i^2$}, % (top left), 
the $M$ least significant bits (LSBs) are outputted to the compute datapath. At the same time, the basis vector is processed by the \textit{basis vector aligner},
% The aligner 
which performs an $M$-bit left circular shift and writes the rotated vector to the OVSF FIFO.
If {\small $M$$>$$K_i^2$}, % (top right), 
the basis vector is self-concatenated {\small $\left\lfloor \frac{M}{K_i^2} \right\rfloor$} times and written to the output's LS part. Simultaneously, the {\small $\text{mod}(M,K_i^2)$} LSBs of the basis vector are written to the output's MSBs and the constructed vector is passed to the compute datapath. In this case, the aligner performs a left circ-shift of \mbox{{\small $K_i^2$-$\text{mod}(M,K_i^2)$}} %bits 
and writes the result to the FIFO.

With this approach, when the basis vectors are read again out of the OVSF FIFO after $\rho K_i^2$ cycles (\textit{i.e.} in the next iteration of the loop on line 2), they are correctly aligned to directly match TiWGen's tiling pattern. For instance, after the generation of the red-striped subtile in Fig.~\ref{fig:weights_gen_tiling}, the FIFO-read basis vectors will be correctly aligned in order to generate the blue-striped subtile without the need for costly selection logic or redundant storage. 
% The basis vector aligner performs circular shifts with a \textit{fixed} number of bit-shifts. As a result, it is implemented through direct connections rather than expensive multiplexers.
\blue{For CNNs with multiple filter sizes, the basis vector aligner is instantiated with as many circ-shift options. As the distinct filter sizes are known \textit{a priori}, only the required shifting logic is inserted, avoiding expensive generic multiplexers, and the appropriate per-layer bit-shift is selected at run time. }

Overall, the proposed design offers a twofold gain. First, it alleviates the redundant replicated storage of basis vectors and avoids the hardware cost of partitioning multiplexers that would require excessive LUTs usage. Second, it provides the necessary bandwidth to the compute datapath while efficiently utilising the on-chip memory through the OVSF FIFO and the resource-efficient aligner design.
As the values of $K_i^2$ for each layer and $M$ are known at design time and after the DSE phase respectively, the OVSF generator is statically instantiated at compile time with the appropriate design. % between the two alternatives.

\vspace{-0.3em}
\subsection{Input Selective PEs for Reducing Underutilisation}
\label{sec:pe_design}
\vspace{-0.2em}

\begin{figure}[t]
    \centering
    \vspace{-0.2cm}
    % \fbox
    {
    \includegraphics[width=0.6\columnwidth,trim={6.5cm 5cm 10.25cm 1cm},clip]{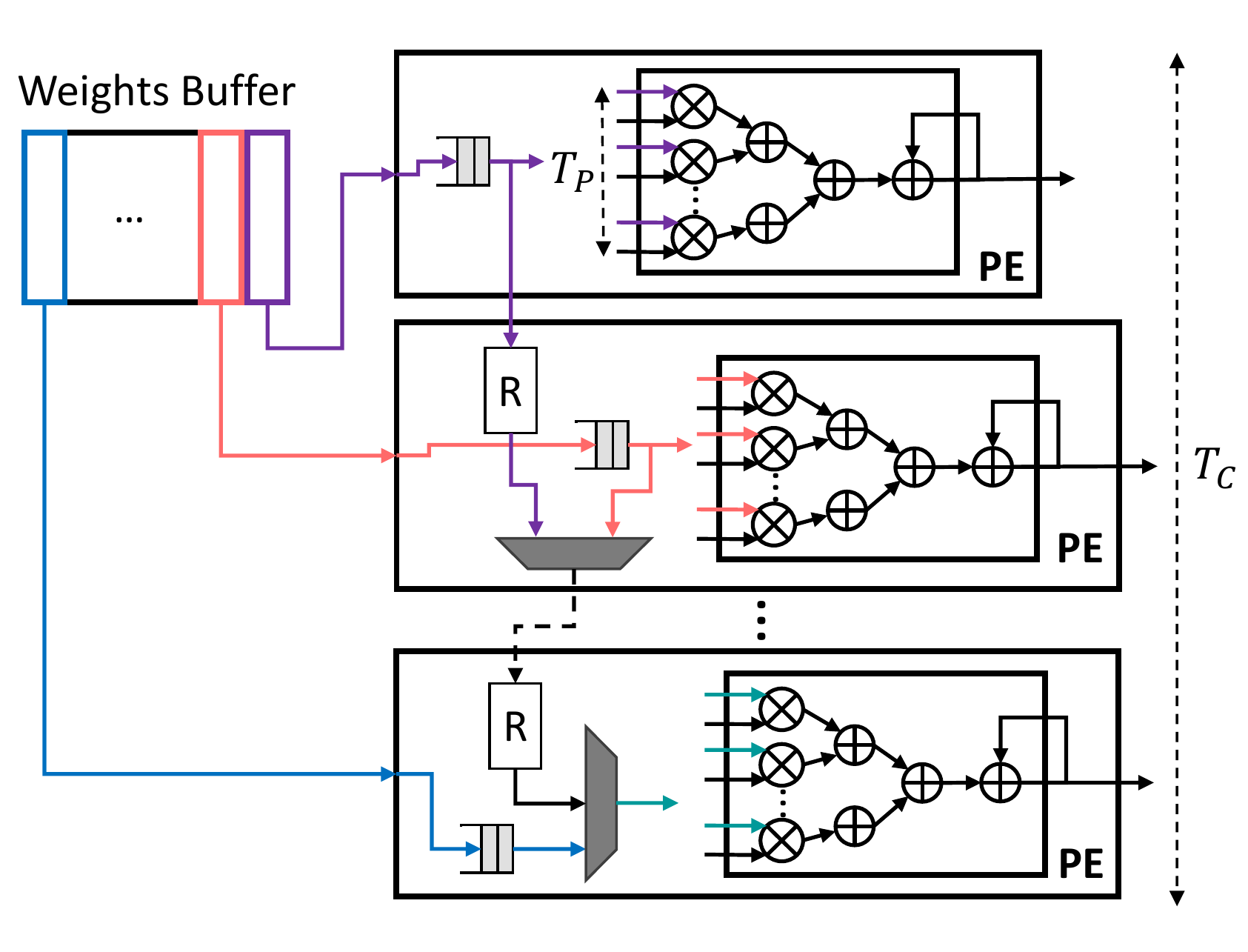}
    }
    \vspace{-0.2cm}
    \caption{\footnotesize \tool's input selective PE array for CNNs.}
    % \vspace{-0.2cm}
    \label{fig:new_pe_design}
\end{figure}

When processing compute-bound layers, existing single computation engines are often bounded by the underutilisation of the computational resources~\cite{latency2017fpl,alamo2020tcad,caffeine2019tcad}. This is due to the fixed CNN engine configuration that does not match all layer dimensions. For instance, mapping a layer with $N_\text{out}$$=$$64$ (\textit{i.e.}~$C$$=$$64$) on an engine with 128 PEs (\textit{i.e.}~$T_C$$=$$128$) would leave 50\% of the PEs idle, halving the attainable performance.  Fig.~\ref{fig:new_pe_design} shows \tool's approach of alleviating this by introducing input selective PEs. The initial PE is augmented with registers and switches. However, not all PEs have the same components; only the PEs that remain underutilised even for a single layer are further equipped with a compact switch that selects the inputs to the dot-product unit. In this manner, the PE can utilise either i)~the default weight written by the weights generator in the weights buffer or ii)~in the absence of this weight, the \blue{weight} passed from the neighbouring PE. In the second case, the weights are propagated along the PE array so that a different weight is used by each augmented PE on each cycle. \blue{Moreover, the Input Buffer (Fig.~\ref{fig:high_level_arch}) is reorganised accordingly to provide parallel access to multiple rows.}

Effectively, this design works as a load-balancing mechanism that partially unrolls the $T_R$ dimension and thus distributes the work more evenly among the PEs. By restricting connectivity only to neighbouring units and enhancing only the underutilised PEs, the additional circuitry is low-overhead and delivers up to 20\% higher performance on compute-bound layers.

\vspace{-0.4em}
\section{Design Space Exploration}
\label{sec:dse}
\vspace{-0.2em}

To estimate the performance and resource usage of different architectural parameters, 
an analytical modelling framework was developed. 
All models have been verified empirically.

\textbf{Performance Model.}
The workload of a CNN with $N_L$ layers is represented as a sequence of {\small $W_i$$=$$\left<R_i,P_i,C_i\right>$} \textit{workload tuples} with $i~\in~\{1,...,N_L\}$.
Given a design point {\small $\sigma$$=$$\left<M,T_R,T_P,T_C\right>$}, the \texttt{CNN-WGen}'s runtime for generating the i-th layer's weights %required 
to compute an %{\small $(T_R\times T_C)$} 
output tile is given by % Eq.~(\ref{eq:wgen_exec_time}), with one factor for each of the pipelined loops in Algorithm~\ref{alg:tiled_weights_gen}. 
\vspace{-0.1cm}
\begin{equation}
    \footnotesize
    \vspace{-1mm}
    \resizebox{0.7\linewidth}{!}{
    $t_{\texttt{CNN-WGen}}^i(\sigma, W_i) = \left\lfloor \rho \cdot l \right\rfloor \cdot \left\lceil \frac{T_P \cdot T_C}{M} \right\rceil \cdot \left\lceil \frac{P_i}{T_P} \right\rceil$
    }
    \vspace{-0.5mm}
    \label{eq:wgen_exec_time}
\end{equation}
where $\rho$ and $l$ are the OVSF ratio and basis length respectively, and with one factor for each of the pipelined loops in Algorithm~\ref{alg:tiled_weights_gen}.
With the $\alpha$ values transferred upfront and the OVSF method generating all weights on-chip, the off-chip memory transfers involve only the input and output activations:
\begin{equation}
    \footnotesize
    \resizebox{0.45\textwidth}{!}{
    $t_{\text{mem in}}^i(\sigma, W_i) = \frac{T_R \cdot P \cdot WL}{bw_\text{in}}, \quad t_{\text{mem out}}^i(\sigma, W_i) = \frac{T_R \cdot T_C \cdot WL}{bw_\text{out}}$
    }
    \label{eq:transfer_times}
\end{equation}
where $WL$ is the adopted wordlength, and $bw_{\{\text{in},\text{out}\}}$ are the memory bandwidths for transferring inputs/outputs.

With the $T_C$ and $T_P$ dimensions unrolled, the computation of an output tile %by the accelerator's CNN engine 
requires the pipelined processing of $\frac{P_i}{T_P}$ tiles for each of the $T_R$ rows. Hence, the CNN engine's runtime for each output tile is estimated as {\small $t_{\text{eng}}^i(\sigma, W_i) = T_R \left\lceil \frac{P_i}{T_P} \right\rceil$}. By applying the input selective PEs, the runtime is refined as
\begin{equation}
    \footnotesize
    \resizebox{0.45\textwidth}{!}{
    $t_{\text{eng}^*}^i (\sigma, W_i) = \left( T_C-C_i + \left\lceil \frac{T_R \cdot C_i - (T_C-C_i) \cdot (C_i+1)}{T_C} \right\rceil \right) \cdot \left\lceil \frac{P_i}{T_P} \right\rceil$
    }
    \label{eq:cnn_engine_exec_time_enhanced}
\end{equation}
where the $T_R$ dimension is partially unrolled by processing different rows of $T_R$ through the underutilised PEs.

Overall, the accelerator forms a pipeline of three coarse stages:
the concurrent input transfer and weights generation, the CNN engine processing and the output transfer. 
In this context, the initiation interval of the architecture is given by
\vspace{-1mm}
\begin{equation}
    \footnotesize
    \vspace{-1mm}
    \resizebox{0.91\linewidth}{!}{
    $II^i(\sigma, W_i) = \max\left( \max\left(t_{\text{mem in}}^i, t_{\texttt{CNN-WGen}}^i \right), t_{\text{eng}^*}^i, t_{\text{mem out}}^i \right)$
    }
    \label{eq:initiation_interval}
\end{equation}
% 
%The total runtime for layer $i$ is given by 
and \mbox{{\footnotesize $t_\text{total}^i(\sigma, W_i)=II^i(\sigma, W_i) \left\lceil \frac{R_i}{T_R} \right\rceil \left\lceil \frac{C_i}{T_C} \right\rceil$}} yields the total runtime for the i-th layer.
Thus, given a CNN's workload tuple {\footnotesize $W$$=$$\left<W_i ~|~ \forall i \in \{1, ..., N_L \} \right>$}, the throughput in inferences per sec (inf/s) is estimated as \mbox{{\footnotesize $T(\sigma, W) = 1/\sum\limits_{i=1}^{N_L} t_\text{total}^i (\sigma, W_i)$}}.

\textbf{Resource Consumption Model.}
From a resource perspective, the main design constraints are the DSPs and on-chip RAM blocks of the target FPGA. Assuming that all MAC operators are mapped to DSPs, the values of {\small $
\left<M,T_P,T_C\right>$} are constrained as {\footnotesize $D_\text{MAC} \times (M + T_P T_C) \leq D_\text{fpga}$},
with {\footnotesize $D_{\text{fpga}}$} the available DSPs and {\footnotesize $D_\text{MAC}$} the DSPs/MAC. In our case, 16-bit fixed-point precision is used, where {\footnotesize $D_\text{MAC}$$=$$1$} on the evaluated Xilinx FPGAs.

In terms of on-chip RAM, the accelerator has the I/O and Alpha buffers with wordlength $WL$ and the binary OVSF FIFO, with a total capacity requirement as given by Eq.~(\ref{eq:onchip_ram}).
\vspace{-1mm}
\begin{equation}
    % \footnotesize
    \vspace{-1mm}
    \resizebox{0.91\linewidth}{!}{
    $\left(2(T_R T_P + T_R T_C) + D^{\text{Alpha}}N_P^{\text{Alpha}}\right) WL + K_\text{max}^2K_\text{max}^2 \leq C_\text{fpga}$
    }
    \label{eq:onchip_ram}
\end{equation}
where the factor of 2 accounts for double-buffering and $C_\text{fpga}$ is the on-chip RAM capacity of the target device.

To further estimate the consumption of LUTs, we used a set of place-and-route measurements and fitted linear regression models as a function of \tool's tunable parameters.
Overall, we formally capture the %on-chip 
resource consumption of a design point $\sigma$ by means of vector $\textbf{\textit{rsc}}(\sigma)$ that holds the utilised amount of DSPs, BRAMs and LUTs. Similarly, we denote the resource vector of the target platform by $\textbf{\textit{rsc}}_\text{Avail.}$. 

\textbf{Configuration Optimisation.}
To yield the highest performing design for the given CNN-FPGA pair, we cast DSE as a formal optimisation problem: \mbox{{\small $\max\limits_{\sigma} T\left(\sigma, W \right) ~ \text{s.t.} ~ \textbf{\textit{rsc}}(\sigma) \leq \textbf{\textit{rsc}}_\text{Avail.}$}}. 
Given a CNN-FPGA pair, we perform exhaustive search, exploring different resource allocations between \texttt{CNN-WGen} and the CNN engine. All designs that violate the resource constraints are pruned as infeasible to accelerate the exploration.

\vspace{-0.4em}
\section{Lightweight OVSF models}
\label{sec:OVSFmodels}

To efficiently derive lightweight OVSF models from a CNN, we introduce two techniques: 1)~a regression scheme that enables the use of readily available pretrained models, reducing the costly training on large datasets; and 2)~two mechanisms to overcome the limitation of OVSF codes not being able to directly represent $3$$\times$$3$ filters, with marginal accuracy drop.

\red{Convolutional layers in OVSF models learn weighting coefficients of} OVSF \red{codes} that are linearly combined to generate the filters. To exploit the availability of pretrained models, we obtain the scalars, $\alpha_{i}$, from a pretrained filter $\hat{f}_{i}$ by framing the minimisation problem as a regression:
\vspace{-2mm}
\begin{equation}
    \scriptsize
    \vspace{-1mm}
    \alpha^{*}_{i} = \argmin_{\alpha_{i}}\left\|f_{i} - \hat{f}_{i} \right \|_2^2 = \argmin_{\alpha_{i}} \left \| \sum_{j = 0}^{\lfloor \rho \cdot l \rfloor}\alpha_{i}^{j}B_{i}^{j} - \hat{f}_{i} \right \|^2_2
    \label{eq:regression}
\end{equation}
When ratio $\rho$$=$$1$ and shape of filter $f_{i}$ matches the OVSF condition of having $2^{n}$ with $n$$\in$$\mathbb{N}$ parameters, then this minimisation guarantees $f_{i}$ being indistinguishable from $\hat{f}_{i}$. \red{When} $\rho$$<$$1$, we follow an iterative approach to greedily discard OVSF bases, based on their scalars' magnitude, until the specified $\rho$ is reached. The regression stage comprises a 2-layer MLP network and 100 iterations per filter.

We evaluated two approaches to overcome \red{the limitation of generating $3$$\times$$3$ filters using OVSF codes}: first, by learning a $3$$\times$$3$ crop from a $4$$\times$$4$ filter (which can be represented using OVSF codes); secondly, by means of an auxiliary layer that performs adaptive average pooling transforming a $4$$\times$$4$ filter into the desired $3$$\times$$3$ shape. We empirically found that taking a $3$$\times$$3$ crop worked best \blue{for all models except ResNet50}.

\vspace{-0.6em}
\section{Evaluation}
\label{sec:evaluation}

\vspace{-0.3em}
\subsection{Experimental Setup}
\label{sec:exp_setup}
\vspace{-0.2em}

We target two FPGA platforms with different resource characteristics:
ZC706 mounting the mid-tier Z7045 and ZCU104 with the larger ZU7EV, with a clock rate of 150 and 200 MHz respectively. Our hardware designs were synthesised and placed-and-routed with Xilinx Vivado HLS and Vivado Design Suite (v2019.2) and run on both boards. The corresponding Arm CPU was used to set up the off-chip memory transactions, launch %the hardware 
execution and measure the end-to-end performance of each design. In the evaluation, 16-bit fixed-point precision was used, with \tool also supporting other quantisation schemes. The available off-chip memory bandwidth was controlled by using a different number of memory ports and amount of word packing, spanning from 1.1 GB/s (1$\times$) to 13.4 GB/s (12$\times$).

\textbf{Benchmarks.}
We evaluate on CNNs of varying depth, workload and memory footprint. In particular, we target the widely used family of residual networks~\cite{DBLP:journals/corr/HeZRS15}. Concretely, we use ResNet18, ResNet34 and \blue{ResNet50} on the ImageNet dataset. 
We also evaluate \tool on  
SqueezeNet1.1~\cite{iandola2016squeezenet}, a highly optimised network for resource-constrained devices.

\textbf{Training Scheme.}
We have developed \tool's training component on top of \textit{PyTorch} (1.5)~\cite{NIPS2019_9015}.
To derive the OVSF models, we modified the official \textit{PyTorch}-based ResNet by replacing all $3$$\times$$3$ convolutional layers within residual blocks with their OVSF counterparts. In all our experiments, we employed pretrained ImageNet models from \textit{torchvision} (0.6.0). After a regression stage that transforms standard models into OVSF ones, the models were fine-tuned for 30 epochs using an Adam optimiser~\cite{kingma2014adam} and learning rate decay every 10 epochs. For each given model, we trained two OVSF variants following different \red{distributions} of ratios $\rho$ for
%each $3$$\times$$3$ convolutional layer 
\red{layers} in each of the four residual blocks. First, OVSF50 with ratios=$[1.0,0.5,0.5,0.5]$; and OVSF25 with ratios=$[1.0,0.4,0.25,0.125]$. 
We follow the same procedure and ratios for SqueezeNet's \textit{Fire} modules.

\begin{table}[t]
    \tablefontsize
    \centering
    \captionsetup{font=small,labelfont=bf}
    \caption{Accuracy and number of parameters for ResNet34 models on ImageNet following different compression schemes. Performance measured on the Zynq 7045 platform at different memory bandwidths.}
    \vspace{-0.1cm}
    \resizebox{1.0\columnwidth}{!}{
    \begin{tabular}{l c c c c}
        \toprule
        Model & Compression & Params & Accuracy & Performance (inf/sec) \\
        Arch. & Method & (millions) & (\%) & ($1\times$, $2\times$, $4\times$) \\%, $12\times$)\\
        \midrule
        ResNet34 & - & 21.8 & 73.3 & (8.6, 16.8, 28.7) \\ %, 36.0)\\
        \midrule
        ResNet34 & Tay82 & 17.4 & $72.7$ & (10.7, 21.0, 35.6) \\ %, 44.3) \\
        ResNet34 & Tay72 & 15.1 & $71.9$ & (13.3, 25.8, 44.0) \\ %, 54.7) \\
        ResNet34 & Tay56 & 9.4 & $67.8$ & (18.3, 36.3, 63.8) \\ %, 77.6) \\
        ResNet34 & Tay45 & 6.3 & $63.1$ & (21.8, 43.4, 79.8) \\ %, 97.9) \\
        \midrule
        ResNet34 & OVSF50 & 17.2 & $72.8$ & (18.1, 21.8, 31.1) \\ %, 33.3) \\
        ResNet34 & OVSF25 & 7.2 & $71.5$ & (18.4, 27.3, 33.5) \\ %, 33.7) \\
        \midrule
        ResNet34 & Tay82+OVSF50 & 13.2 & $71.1$ & (18.6, 30.0, 37.3) \\ %, 38.3) \\
        ResNet34 & Tay82+OVSF25 & 6.7 & $70.6$ & (18.8, 31.0, 38.9 )\\ %, 40.5) \\
        ResNet34 & Tay72+OVSF50 & 11.9 & $70.3$ & (18.8, 32.0, 40.2) \\ %, 41.3) \\
        ResNet34 & Tay72+OVSF25 & 4.9 & $68.9$ & (18.9, 33.3, 42.0) \\ %, 43.3) \\
        \bottomrule
    \end{tabular}
    }
    \label{tab:accResultsResnet34}
\end{table}

\textbf{Baselines.}
We introduce two highly optimised single computation engines executing: a)~the vanilla CNN and b)~pruned variants. For b), we use a state-of-the-art method~\cite{Molchanov_2019} which applies channel pruning based on the first-order Taylor approximation contribution of each filter to the model's loss. This process is carried out iteratively until a target compression ratio is reached. We refer to a pruned model that keeps 82\% of the filters as Tay82 and follow the same naming scheme for other ratios.
The baseline architecture comprises the same design as \tool with the weights transferred from the off-chip memory into an additional {\small $T_P$$\times$$T_C$} buffer, if they do not fit on-chip. Both a) and b) are parametrised with tile sizes {\small $\left<T_R,T_P,T_C \right>$} and roofline modelling~\cite{cnnroofline2015fpga} is used to obtain the highest throughput configuration for the target \mbox{CNN-FPGA pair}. 

\vspace{-0.3em}
\subsection{Performance Comparison}
\label{sec:perf_comparison}
\vspace{-0.2em}

This section assesses the actual performance gains of \tool compared to our optimised baselines.
Tables~\ref{tab:accResultsResnet34} and~\ref{tab:accResultsResnet18} show the achieved validation set accuracy and actual performance of each design as measured on ZC706 under varying bandwidth budget. Across bandwidths (1$\times$/2$\times$/4$\times$ where 4$\times$ is the 4.5 GB/s peak measured bandwidth on ZC706), \tool's OVSF50 and OVSF25 designs outperform the faithful baseline by {\small 2.1$\times$/1.3$\times$/1.1$\times$} and {\small 2.1$\times$/1.6$\times$/1.2$\times$} respectively for ResNet34, and by {\small 1.6$\times$/1.6$\times$/1.24$\times$} and {\small 1.4$\times$/1.5$\times$/1.3$\times$} respectively for ResNet18. As bandwidth availability increases, the baseline becomes less memory-bound and the performance gap closes. 
Table~\ref{tab:accResultsSqueezenet} shows the comparison of \tool with the faithful baseline for SqueezeNet on ZU7EV with peak measured bandwidth of 13.4~GB/s (12$\times$). Both OVSF50 and OVSF25 designs yield increasing throughput gains as the bandwidth becomes more restricted, with OVSF25 sustaining over 57\% speedup for up to 4$\times$ bandwidth.
\blue{
Under 1$\times$ bandwidth, OVSF25 offers minimal additional gains. This is because, below a compression ratio, even though the memory needs are further reduced, activations begin to dominate I/O, and hence further weights reduction does not provide significant benefits. Activations compression techniques~\cite{eyeriss2017jssc,scnn2017isca} can be orthogonally combined to obtain further gains.
}

\begin{table}[t]
    \tablefontsize
    \centering
    \captionsetup{font=small,labelfont=bf}
    \caption{Accuracy and number of parameters for ResNet18 models on ImageNet following different compression schemes. Performance measured on the Zynq 7045 platform at different memory bandwidths.}
    \vspace{-0.1cm}
    \resizebox{1.0\columnwidth}{!}{
    \begin{tabular}{l c c c c}
        \toprule
        Model & Compression & Params & Accuracy & Performance (inf/sec) \\
        Arch. & Method & (millions) & (\%) & ($1\times$, $2\times$, $4\times$) \\ %, $12\times$)\\
        \midrule
        ResNet18 & - & 11.7 & 69.8  & (12.0, 23.5, 40.1)\\ %, 54.5) \\
        \midrule
        ResNet18 & Tay88 & 9.1 & $68.8$ & (14.3, 28.0, 46.4)\\ %, 61.9) \\
        ResNet18 & Tay82 & 7.9 & $67.3$ & (14.3, 27.8, 45.4)\\ %, 61.5) \\
        ResNet18 & Tay72 & 6.0 & $64.8$ & (18.2, 35.3, 57.6)\\ %, 77.3) \\
        ResNet18 & Tay56 & 3.7 & $58.3$ & (23.8, 47.3, 82.2)\\ %, 99.5) \\
        \midrule
        ResNet18 & OVSF50 & 9.1 & $69.2$ & (19.4, 33.8, 49.9)\\ %, 51.9) \\
        ResNet18 & OVSF25 & 4.1 & $67.3$ & (19.4, 34.8, 51.0)\\ %, 52.7) \\
        \midrule
        ResNet18 & Tay82+OVSF50 & 6.3 & $66.2$ & (24.5, 43.2, 57.9)\\ %, 58.5) \\
        ResNet18 & Tay82+OVSF25 & 2.8 & $64.4$ & (24.5, 43.6, 59.7)\\ %, 61.2) \\
        \bottomrule
    \end{tabular}
    }
    \label{tab:accResultsResnet18}
    % \vspace{-0.1cm}
\end{table}

Compared to the pruned baselines, \tool's OVSF models are more resilient at high compression ratios while resulting in similar accuracy at lower compression ratios. In terms of throughput, \tool delivers faster processing at more constrained bandwidths. Concretely, ResNet34-OVSF50 is 80\% faster than Tay82 at 1$\times$ bandwidth, with less than 1 percentage point (pp) accuracy drop. 
% Despite both models having almost identical model sizes, ResNet34-OVSF50 results in less memory-bound layers across the network, contributing to a significantly higher throughput over Tay82 at low bandwidths.
% Despite both models having almost identical sizes, \blue{Tay82 achieves its reduction by pruning mostly compute-bound layers, whereas memory-bound layers still throttle the performance. ResNet34-OVSF50 compresses more effectively the memory-bound layers, thus leading to significantly higher throughput at low bandwidths.} 
% \javier{I'm unsure if we should make this kind of distinction. Taylor compresses layers based on their contribution to the loss, and OVSF is done statically based on those per-block manually-selected ratios. I'm not sure if these conclusions about OVSF being more effective in memory-bound layers would raise questions about how fairly we compared each technique.}
Despite \red{being almost identical in terms of model size and accuracy}, Tay82's approach, which prioritises the pruning of layers with the least accuracy impact, leads to the pruning of mostly compute-bound layers when targeting ResNet34. On the other hand, ResNet34-OVSF50 compresses more effectively memory-bound layers, leading to significantly higher throughput at low bandwidths.
A similar pattern is observed for ResNet18. At higher compression ratios, ResNet34-OVSF25 yields 3.7 pp higher accuracy than Tay56, despite using 25\% fewer parameters.

To explore the benefits of combining \tool's OVSF execution scheme with channel pruning, we derive, train and map on \tool Tay-OVSF models. 
This approach yields competitive lightweight models that are not attainable through pruning alone. For instance, ResNet18 with Tay82+OVSF25 is 25\% smaller than ResNet18-Tay56 and achieves 6.1 pp higher accuracy, while achieving 34.6\% and 23.5\% higher throughput over ResNet18-Tay72 with less than 0.5 pp accuracy drop. 

\begin{table}[t]
    % \vspace{-0.2cm}
    \small
    \centering
    \captionsetup{font=small,labelfont=bf}
    \caption{Comparing \tool with faithful baseline on SqueezeNet on ImageNet. Performance measured on the UltraScale+ ZU7EV platform at different memory bandwidths.}
    \vspace{-0.1cm}
    \resizebox{1.0\columnwidth}{!}{
    \begin{tabular}{l c c c c}
        \toprule
        Model & Compression & Params & Accuracy & Performance (inf/sec) \\
        Arch. & Method & (millions) & (\%) & ($1\times$, $2\times$, $4\times$, $12\times$)\\
        \midrule
        SqueezeNet & - & 1.24 & 58.2 & (54.7, 108.9, 217.8, 515.6) \\
        \midrule
        SqueezeNet & OVSF50 & 1.07 & 57.6 & (97.4, 189.7, 339.1, 594.1) \\
        SqueezeNet & OVSF25 & 0.86 & 57.1 & (97.4, 189.7, 342.6, 600.5) \\
        \bottomrule
    \end{tabular}
    }
    \label{tab:accResultsSqueezenet}
\end{table}

\blue{
\textbf{Comparison with existing FPGA designs.}
Here, we evaluate \tool with diverse existing FPGA designs. In all cases, we use the OVSF50 variant with less than 1-pp accuracy drop.
% Table~\ref{tab:comparison_table} presents the comparison for ResNet18/34 and SqueezeNet. 
As shown in Table~\ref{tab:comparison_table}, we compare with Snowflake~\cite{snowflake2017compiling} for ResNet18, the SOTA sparse CNN FPGA design~\cite{sparsecnnaccel2019fccm} for ResNet34, and the light-CNN-optimised Light-OPU~\cite{lightopu2020fpga} and \cite{maximising2017isca} that addresses PE underutilisation for SqueezeNet. On Z7045, \tool achieves 2.33$\times$ and 1.12$\times$ higher throughput than \cite{snowflake2017compiling} and \cite{sparsecnnaccel2019fccm}, respectively. For SqueezeNet, our design delivers 1.83$\times$ and 1.67$\times$ higher performance density in inf/s/DSP and inf/s/Logic, respectively, than Light-OPU and 1.40$\times$ in inf/s/DSP and 1.14$\times$-1.27$\times$ in inf/s/Logic over \cite{maximising2017isca} with the same or 36\% less on-chip memory.
}

\blue{
\textbf{ResNet50 Results.}
The original ResNet50 reaches 76.15\% accuracy with 25.56M parameters. Our ResNet50-OVSF50 improves accuracy to 76.23\% with 22.83M parameters. 
Table~\ref{tab:resnet50_comparison_table} presents the comparison for ResNet50. On Z7045, \tool outperforms Snowflake by 1.59$\times$ in inf/s. Compared with designs on larger devices, our design achieves higher performance density (GOp/s/DSP) by 3.71$\times$, 3.69$\times$, 1.29$\times$ and 1.76$\times$-6.16$\times$ over ResNetAccel, xDNN, DNNVM and ALAMO, respectively, and consistently higher GOp/s/Logic.
FTDL reaches higher GOp/s/DSP and 1.47$\times$ lower GOp/s/Logic, but targets a platform with 2.33$\times$ larger on-chip memory and 2$\times$ higher bandwidth, which substantially reduce the off-chip memory accesses and the associated latency.
}

\textbf{Resource Usage.}
We select \tool and baseline designs with up to 1-pp accuracy drop and compare their post place-and-route resource usage on Z7045, reported in [DSPs, BRAM, LUTs] tuples for 4$\times$ bandwidth.
For ResNet34, the faithful baseline consumes {\small [99\%,83\%,77\%]}, Tay82 {\small [99\%,79\%,77\%]}, OVSF50 {\small [100\%,81\%,78\%]} and Tay82+OVSF50 {\small [100\%,87\%,81\%]}. For ResNet18, the faithful baseline {\small [78\%,99\%,70\%]}, Tay88 {\small [78\%,99\%,69\%]}, OVSF50 {\small [100\%,87\%,75\%]} and Tay82+OVSF50 {\small [100\%,83\%,80\%]}. \blue{For ResNet50, OVSF50 on ZU7EV consumes {\small [100\%,87\%,78\%]}.} Finally, the input selective PE mechanism adds a minimal LUTs overhead of less than 7\%.

\begin{table*}[t]
	\centering
	\captionsetup{font=small,labelfont=bf}
    \caption{Comparison with existing FPGA work on ResNet18 (4.03 GOps), ResNet34 (7.40 GOps) and SqueezeNet (0.78 GOps).}
	\vspace{-0.2cm}
	\resizebox{1\textwidth}{!}{
	\setlength\tabcolsep{10pt} % default value: 6pt
		\begin{threeparttable}
			%\vspace{-0.1cm}
			\small
			%	\footnotesize
			%	\normalsize
			%	\huge
			%	\large
			%\resizebox{0.5\textwidth}{!}{
			%	\resizebox{\linewidth}{!}{
			%	\resizebox{0.5\textwidth}{!}{
			\begin{tabular}{l l l| l l| l l l l}
				%\hline
				\toprule
				& ResNet18~\cite{snowflake2017compiling} 
				& \begin{tabular}[l]{@{}l@{}} \tool: \\ ResNet18* \end{tabular}
				%& ResNet34~\cite{sparsecnnaccel2019fccm} 
				& \begin{tabular}[l]{@{}l@{}} Sparse ResNet34~\cite{sparsecnnaccel2019fccm} \\ using Deep Compression  \end{tabular}
				& \begin{tabular}[l]{@{}l@{}} \tool: \\ ResNet34* \end{tabular} 
				& SqueezeNet~\cite{lightopu2020fpga} 
				& \multicolumn{2}{l}{SqueezeNet~\cite{maximising2017isca}} 
				& \begin{tabular}[l]{@{}l@{}} \tool:\\ SqueezeNet* \end{tabular}
				\\
				\cmidrule{7-8}
				\midrule
				FPGA  
				& Z7045 
				& Z7045 
				& Z7045 
				& Z7045
				& K325T
				% & \begin{tabular}[l]{@{}l@{}} Kintex-7 \\ 325T \end{tabular}
				& V485T 
				% & \begin{tabular}[l]{@{}l@{}} Virtex-7 \\ 485T \end{tabular}
				& V690T
				% & \begin{tabular}[l]{@{}l@{}} Virtex-7 \\ 690T \end{tabular} 
				& ZU7EV 
				\\
				Clock (MHz) & 250 & 150 & 166	& 150 & 200 & 170 & 170 & 200 \\
				Precision & 16b fixed & 16b fixed & 16b fixed & 16b fixed & 8b fixed & 16b fixed & 16b fixed & 16b fixed \\
				DSPs$^\dagger$ & 900 & 900 & 900 & 900 & 840 & 2800 & 3600 & 1728 \\
				Logic Capacity & 218.6 kLUTs & 218.6 kLUTs & 218.6 kLUTs & 218.6 kLUTs & 203.8 kLUTs & 303.6 kLUTs & 433.2 kLUTs & 230.0 kLUTs \\
				%	Fixed-point DSPs* & 220 & 900 & 220 & 900 \\
				Block RAM & 2.40 MB & 2.40 MB & 2.40 MB & 2.40 MB & 1.95 MB & 4.52 MB & 6.46 MB & 4.75 MB \\
				{\color{black}DSP Util.$^\dagger$} & 28.4\% & 100\% & 56.8\% & 100\% & 83.8\% & 80\% & 80\% & 100\% \\
				
				\begin{tabular}[t]{@{}l@{}} 
					inf/s
				\end{tabular} 
				& 21.38
				& 49.90
				& 27.84
				& 31.1
				& 420.90
				& 913.40
				& 1173.00
				& 792.20
				\\
				\begin{tabular}[t]{@{}l@{}} 
					inf/s/DSP$^\dagger$
				\end{tabular} 
				& 0.0237
				& 0.0576
				& 0.0309
				& 0.0369
				& 0.2505
				& 0.3260
				& 0.3258
				& 0.4584
				\\
				\begin{tabular}[t]{@{}l@{}} 
					inf/s/Logic
				\end{tabular} 
				& 0.0978
				& 0.2282 % 0.2372
				& 0.1273
				& 0.1422 % 0.1521
				& 2.0652
				& 3.0085
				& 2.7077
				& 3.444
				\\
			
				\bottomrule
				
			\end{tabular} 
			\begin{tablenotes}
				\footnotesize
				\item * using OVSF50, ** batch size = 1, $\dagger$ 18$\times$18, 19$\times$18 and 25$\times$18 DSP configurations, inf/s/DSP is adjusted based on precision for fair comparison (multiplied by 0.5 for 8b).
			\end{tablenotes}
			
		\end{threeparttable}
	}
	\vspace{-0.25cm}
	\label{tab:comparison_table}
\end{table*}

\begin{table*}[t]
	\centering
	\captionsetup{font=small,labelfont=bf}
    \caption{Comparison with existing FPGA work on ResNet50 (8.41 GOps).}
	\vspace{-0.2cm}
	\resizebox{1\textwidth}{!}{
	\setlength\tabcolsep{8pt} % default value: 6pt
		\begin{threeparttable}
			%\vspace{-0.1cm}
			\small
			%	\footnotesize
			%	\normalsize
			%	\huge
			%	\large
			%\resizebox{0.5\textwidth}{!}{
			%	\resizebox{\linewidth}{!}{
			%	\resizebox{0.5\textwidth}{!}{
			\begin{tabular}{l l l | l l l l l l l}
				%\hline
				\toprule
				& Snowflake~\cite{snowflake2017iscas} 
				& \begin{tabular}[l]{@{}l@{}} \tool: \\ ResNet50* \end{tabular}
				& ResNetAccel~\cite{residaccel2017iscas} 
				& xDNN~\cite{xdnn2020xilinx}
				& DNNVM~\cite{dnnvm2019tcad}
				& \multicolumn{2}{l}{ALAMO~\cite{alamo2020tcad}}
				& FTDL~\cite{ftdl2020dac}
				& \begin{tabular}[l]{@{}l@{}} \tool:\\ ResNet50* \end{tabular}
				\\
				\cmidrule{7-8}
				\midrule
				FPGA  
				& Z7045 
				& Z7045 
				& Arria 10 GX1150 
				& VU9P 
				& ZU9
				& Arria10 GX1150
				% & \begin{tabular}[l]{@{}l@{}} Kintex-7 \\ 325T \end{tabular}
				& Stratix10 GX2800
				% & \begin{tabular}[l]{@{}l@{}} Virtex-7 \\ 485T \end{tabular}
				& VU125
				% & \begin{tabular}[l]{@{}l@{}} Virtex-7 \\ 690T \end{tabular} 
				& ZU7EV 
				\\
				Clock (MHz) & 250 & 150 & 150 & 500 & 500 & 240 & 300 & 650 & 200 \\
				Precision & 16b fixed & 16b fixed & 16b fixed & 8b fixed & 8b fixed & 16b fixed & 16b fixed & 16b fixed & 16b fixed \\
				DSPs$^\dagger$ & 900 & 900 & 3036 & 6840 & 2520 & 3036 & 11,520 & 1200 & 1728 \\
				Logic Capacity & 218.6 kLUTs & 218.6 kLUTs & 427.2 kALMs & 1182.0 kLUTs & 274.0 kLUTs & 427.2 kALMs & 933.0 kALMs & 716.0 kLUTs & 230.0 kLUTs \\
				%	Fixed-point DSPs* & 220 & 900 & 220 & 900 \\
				Block RAM & 2.40 MB & 2.40 MB & 6.60 MB & 9.48 MB & 4.01 MB & 6.60 MB & 28.62 MB & 11.075 MB & 4.75 MB \\
				{\color{black}DSP Util.$^\dagger$} & 28.4\% & 100\% & 56.8\% & 100\% & 83.8\% & 80\% & 80\% & 100\% & 100\% \\
				
				\begin{tabular}[t]{@{}l@{}} 
					inf/s
				\end{tabular} 
				& 17.7
				& 28.18
				& 33.93
				& 153.57
				& 80.95
				& 71.38
				& 77.55
				& 151.22
				& 71.71
				\\
				\begin{tabular}[t]{@{}l@{}} 
					inf/s/DSP$^\dagger$
				\end{tabular} 
				& 0.0196
				& 0.0313
				& 0.0111
				& 0.0112
				& 0.0321
				& 0.0235
				& 0.0067
				& 0.1260
				& 0.0415
				\\
				\begin{tabular}[t]{@{}l@{}} 
					inf/s/Logic
				\end{tabular} 
				& 0.0809
				& 0.1289
				& 0.0794 
				& 0.0649
				& 0.1477 
				& 0.1671
				& 0.0831
				& 0.2112
				& 0.3117
				\\

				\bottomrule
				
			\end{tabular} 
			\begin{tablenotes}
				\small
				\item * using OVSF50, ** batch size = 1, $\dagger$ 18$\times$18, 19$\times$18 and 25$\times$18 DSP configurations, inf/s/DSP is adjusted based on precision for fair comparison (multiplied by 0.5 for 8b).
			\end{tablenotes}
			
		\end{threeparttable}
	}
	\vspace{-0.65cm}
	\label{tab:resnet50_comparison_table}
\end{table*}

\begin{figure}[t]
    % \vspace{-2em}
    \centering
    \begin{subfigure}{0.24\textwidth}
        \centering
        \includegraphics[width=1.25\columnwidth,trim={10cm 4.5cm 6cm 6.2cm},clip]{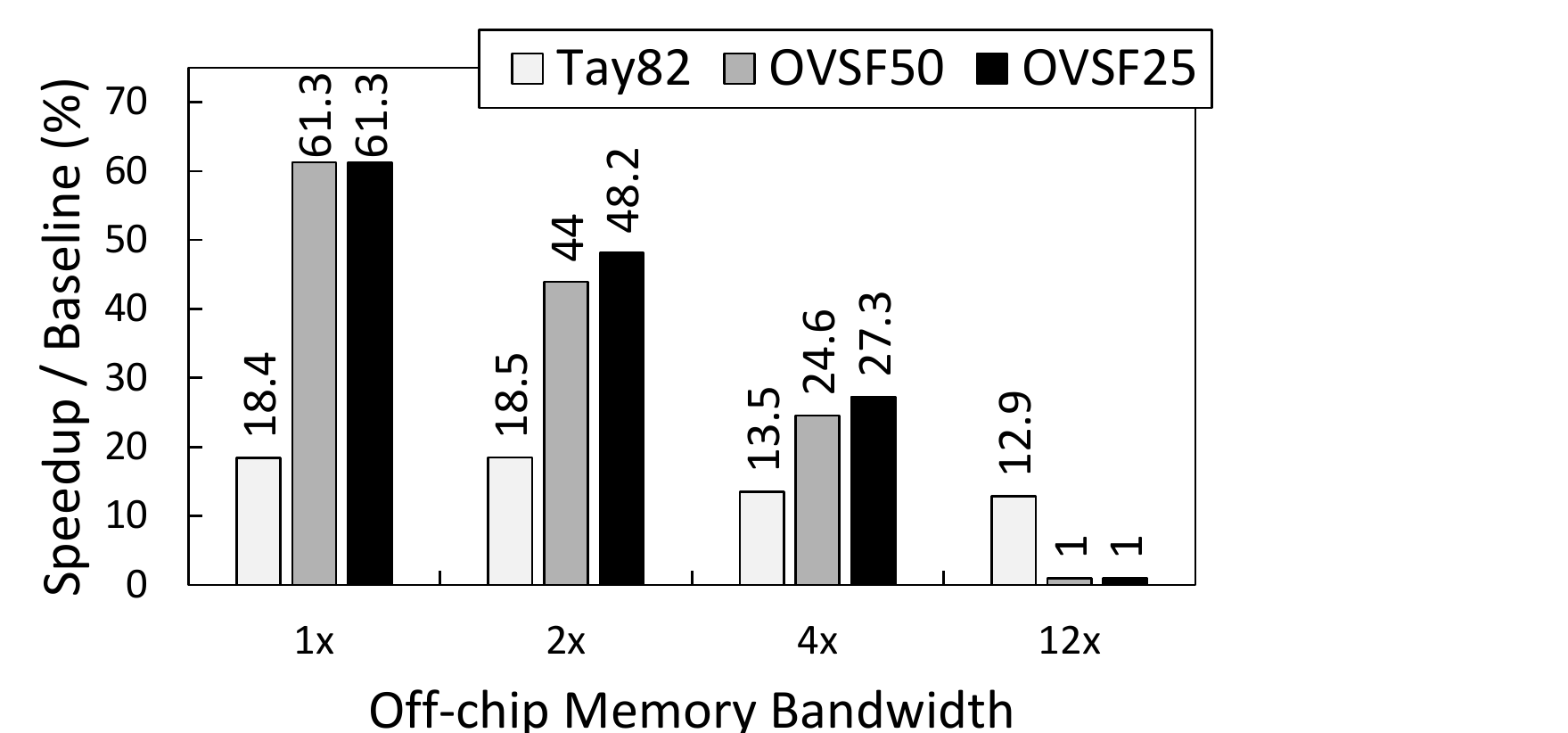}
        \caption{ResNet18 - Z7045}
        \label{fig:resnet18z7045}
    \end{subfigure}
    % \hfill
    \begin{subfigure}{0.24\textwidth}
        \centering
        \includegraphics[width=1.25\columnwidth,trim={10cm 4.5cm 6cm 6.2cm},clip]{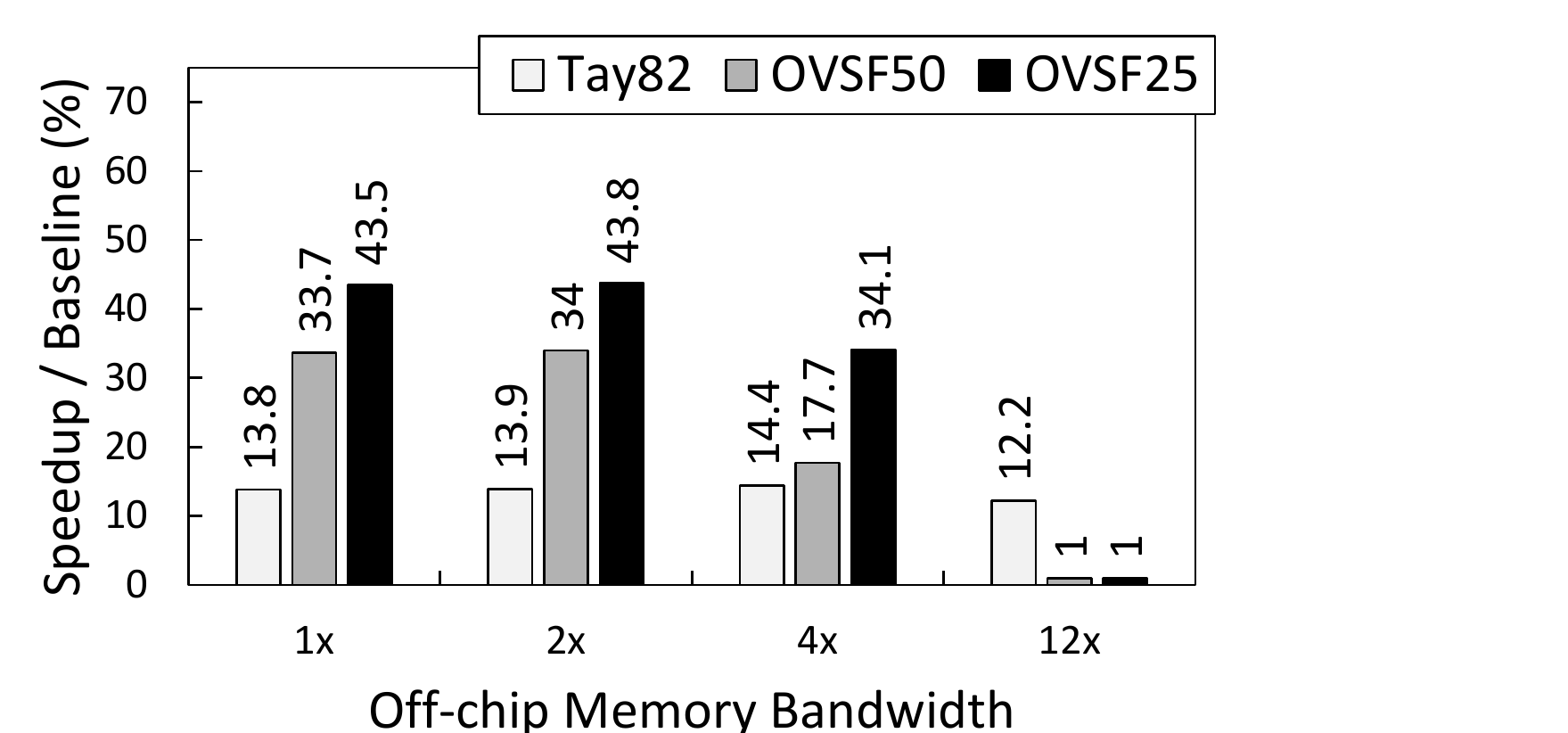}
        \caption{ResNet18 - ZU7EV}
        \label{fig:resnet18zu7ev}
    \end{subfigure}
    % \hfill
    \begin{subfigure}{0.24\textwidth}
        \centering
        % 10cm 5cm 6cm 5cm
        \includegraphics[width=1.25\columnwidth,trim={10cm 4.5cm 6cm 6.2cm},clip]{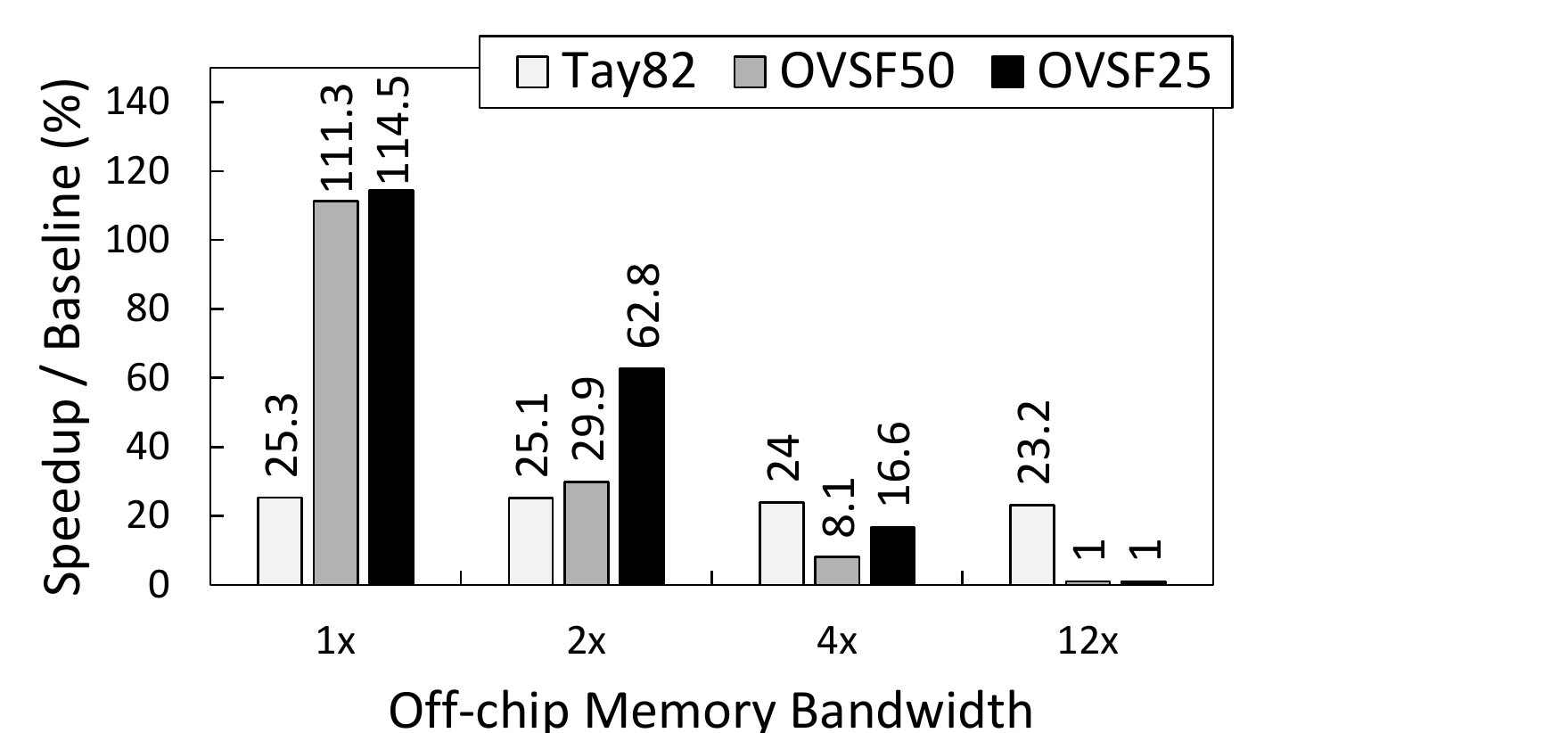}
        \caption{ResNet34 - Z7045}
        \label{fig:resnet34z7045}
    \end{subfigure}
    % \hfill
    \begin{subfigure}{0.24\textwidth}
        \centering
        \includegraphics[width=1.25\columnwidth,trim={10cm 4.5cm 6cm 6.2cm},clip]{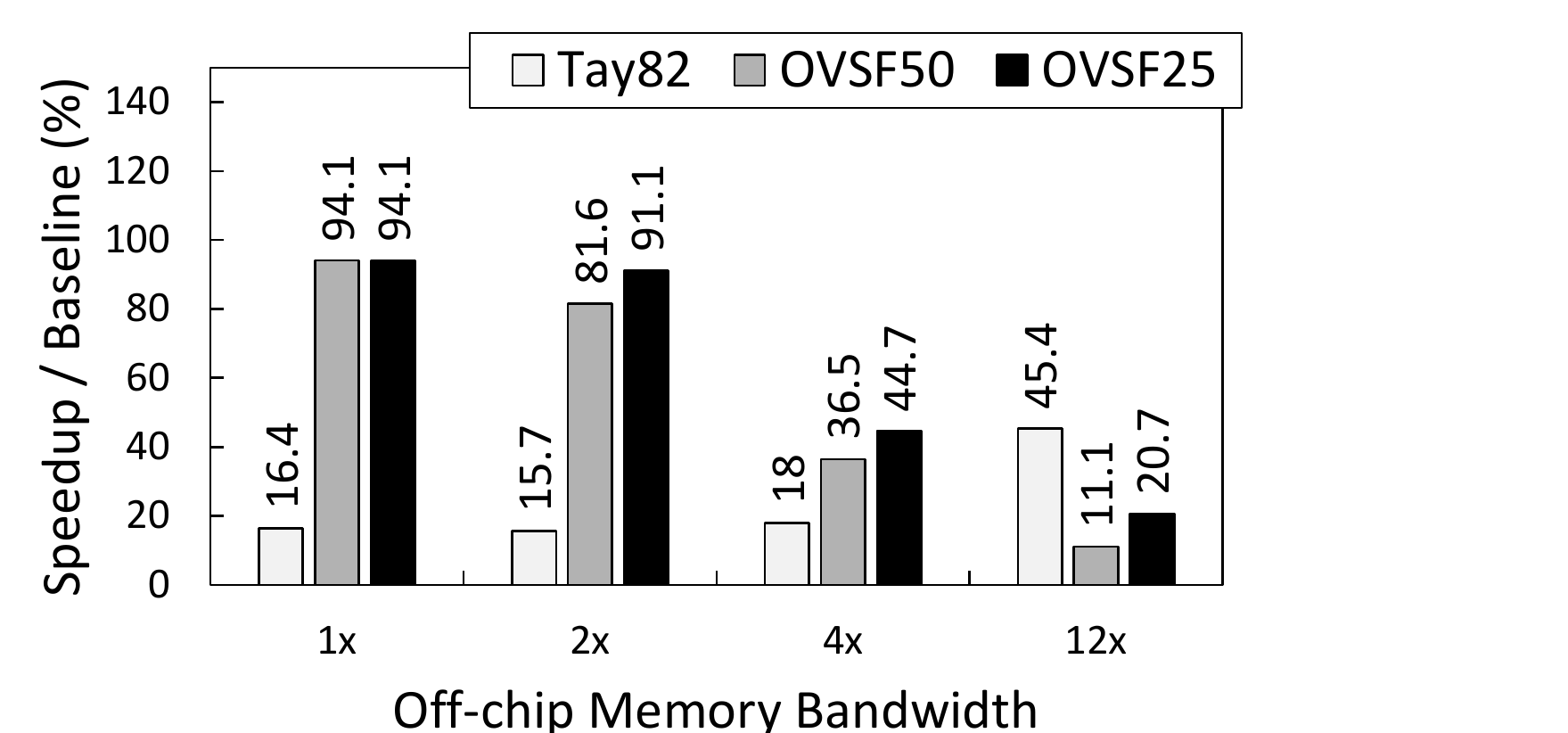}
        \caption{ResNet34 - ZU7EV}
        \label{fig:resnet34zu7ev}
    \end{subfigure}
    \vspace{-0.2cm}
    \caption{\footnotesize Speedup over optimised ResNet18 (a-b) and ResNet34 (c-d) baselines when varying the available off-chip memory bandwidth. 
    }
    \label{fig:bw_sensitivity}
    % \vspace{-0.1cm}
\end{figure}

\vspace{-0.3em}
\subsection{Sensitivity to Off-Chip Memory Bandwidth}
\label{sec:bw_sensitivity}
\vspace{-0.2em}

Fig.~\ref{fig:bw_sensitivity} shows the impact of varying off-chip memory bandwidth over performance on the two target platforms. The figure compares the speedup of \tool and the Tay82 baseline over the vanilla baseline when varying the external memory bandwidth from 1$\times$ to 12$\times$. The bandwidth's impact is most prominent on the larger ZU7EV, where the performance gains are sustained higher across 1$\times$-4$\times$. In the case of the mid-tier Z7045, we observe a sharper drop in the speedup as the bandwidth increases. This is due to the more limited computational resources of Z7045, which makes most CNN layers compute-bound. In contrast, the abundance of computational resources on ZU7EV makes the CNN layers more memory-bound. For instance, at 4$\times$ bandwidth (4.5 GB/s), the vanilla ResNet18 baseline yields DSP utilisation of 71\% on the compute-bound Z7045 and 53\% on the more memory-bound ZU7EV. In this case, \tool significantly improves both cases by mapping ResNet18-OVSF25 with 89\% and 71\% DSP utilisation.
As a result, \tool sustains its gains across a wider range of bandwidths and outperforms Tay82, until the bandwidth-abundant case (12$\times$) where computational resources become the critical factor. In this case, Tay82's lower number of operations due to pruning leads to higher performance.

Across the designs, the input selective PEs contribute an additional speedup of up to 20\%, with varying gains depending on the CNN-FPGA pair and the available bandwidth. For ResNet34-OVSF25 on ZU7EV, disabling this mechanism leads to 0/13.9/3.3/5.9\% lower throughput for the four bandwidths, with a similar pattern observed for the rest of the CNNs. Our input selective PEs effectively improve the performance of suboptimally mapped layers in compute-bound settings, whereas no gain is obtained for the most bandwidth-constrained case (1$\times$) where the designs are severely memory-bound, limiting further improvements through higher PE utilisation.

% \section{Ideas for discussion}

% \todo[inline]{some ideas worth mentioning}
% \begin{itemize}
%     \item mention use cases for 1x and 2x bandwidths: (e.g. concurrent applications using the same FPGA)
%     \item underclock memory to save power.
%     \item as future work: tune OSVF ratios base on whether layer is compute or memory bound.
%     \item ...
% \end{itemize}

\vspace{-0.3em}
\section{Conclusion}
\label{sec:conclusion}
\vspace{-0.2em}
\blfootnote{\vspace{-0.75mm}This work is partially supported by EPSRC grant EP/M50659X/1.}
This paper presents \tool, a CNN inference system that mitigates the limitations of existing FPGA-based CNN engines. By generating weights on-the-fly and selectively balancing the PE work, \tool outperforms both status-quo and pruned CNN engines for the same bandwidth, \blue{while consistently achieving large performance density improvements over state-of-the-art CNN accelerators}. 
The presented system opens up several future paths. Concretely, we envision: the automated tuning of OVSF ratios in a per-layer basis depending on their compute/memory boundedness; 
and enabling the concurrent deployment of multiple CNNs with lower bandwidths, where \tool offers larger speedups over existing designs.

\bibliographystyle{IEEEtran}
\bibliography{references}
\end{document}